\documentclass[lettersize,journal]{IEEEtran}
\usepackage{amsmath,amsfonts}
\usepackage{algorithm} %
\usepackage{algpseudocode} % 
\usepackage{array}
\usepackage[caption=false,font=normalsize,labelfont=sf,textfont=sf]{subfig}
\usepackage{textcomp}
\usepackage{stfloats}
\usepackage{url}
\usepackage{verbatim}
\usepackage{graphicx}
\usepackage{cite}

\usepackage[figuresright]{rotating}
\usepackage{multirow}
\usepackage{float}
\usepackage{booktabs}
\usepackage{threeparttable}
\usepackage{diagbox}
\usepackage[table]{xcolor}

\begin{document}

\title{Adaptive Redundancy Regulation for Balanced Multimodal Information Refinement}

\author{Zhe Yang,
        Wenrui Li, 
        Hongtao Chen,
        Penghong Wang,\\
        Ruiqin Xiong,~\IEEEmembership{~Senior Member,~IEEE}
        Xiaopeng Fan,~\IEEEmembership{~Senior Member,~IEEE}        
\thanks{This work was supported in part by the National Key R\&D Program of China (2023YFA1008501) and the National Natural Science Foundation of China (NSFC) under grant 624B2049 and U22B2035. (Corresponding author: Wenrui Li)}

\thanks{Zhe Yang is with the Department of Computer Science and Technology, Harbin Institute of Technology, Harbin 150001, China, and also with Harbin Institute of Technology Zhengzhou Research Institute, Zhengzhou 450000, China. (e-mail: yzhe610@stu.hit.edu.cn).}

\thanks{Wenrui Li, Penghong Wang, and Xiaopeng Fan are with the Department of Computer Science and Technology, Harbin Institute of Technology, Harbin 150001, China, and also with Harbin Institute of Technology Suzhou Research Institute, Suzhou 215104, China. (e-mail: liwr618@163.com; phwang@hit.edu.cn; fxp@hit.edu.cn).}
\thanks{Hongtao Chen is with the School of Mathematical Sciences, University of Electronic Science and Technology of China, Chengdu, Sichuan 611731, China (e-mail: ht166chen@163.com).}
\thanks{Ruiqin Xiong is with the School of Electronic Engineering and Computer Science, Institute of Digital Media, Peking University, Beijing, 100871, China (e-mail: rqxiong@pku.edu.cn).}
}

\markboth{Journal of \LaTeX\ Class Files,~Vol.~14, No.~10, July~2024}%
{Shell \MakeLowercase{\textit{et al.}}: A Sample Article Using IEEEtran.cls for IEEE Journals}


\maketitle

\begin{abstract}
Multimodal learning aims to improve performance by leveraging data from multiple sources. During joint multimodal training, due to modality bias, the advantaged modality often dominates backpropagation, leading to imbalanced optimization. Existing methods still face two problems: First, the long-term dominance of the dominant modality weakens representation-output coupling in the late stages of training, resulting in the accumulation of redundant information. Second, previous methods often directly and uniformly adjust the gradients of the advantaged modality, ignoring the semantics and directionality between modalities. To address these limitations, we propose Adaptive Redundancy Regulation for Balanced Multimodal Information Refinement (RedReg), which is inspired by information bottleneck principle. Specifically, we construct a redundancy phase monitor that uses a joint criterion of effective gain growth rate and redundancy to trigger intervention only when redundancy is high. Furthermore, we design a co-information gating mechanism to estimate the contribution of the current dominant modality based on cross-modal semantics. When the task primarily relies on a single modality, the suppression term is automatically disabled to preserve modality-specific information. Finally, we project the gradient of the dominant modality onto the orthogonal complement of the joint multimodal gradient subspace and suppress the gradient according to redundancy. Experiments show that our method demonstrates superiority
among current major methods in most scenarios. Ablation experiments verify the effectiveness of our method. The code is available at \url{https://github.com/xia-zhe/RedReg.git}
\end{abstract}


\section{Introduction}
\IEEEPARstart{D}{eep} learning automatically learns representations and decision patterns from massive amounts of data through multi-layer neural networks, achieving breakthrough progress in tasks such as vision, speech, and natural language processing\cite{zhao1,chen1,zhuoyuan1,liu1,zhao2,chen2,zhuoyuan2,liu2}. Multimodal learning integrates complementary information from different modalities such as images, text, speech, and sensors to enhance understanding, generation, and cross-modal reasoning capabilities.\cite{zhao3,chen3,zhuoyuan3,liu3,tang1}. By introducing additional modalities, researchers have improved traditional unimodal tasks and begun to tackle new challenges, such as audio-visual zero-shot learning\cite{STFT,MSTR}, audio-visual question answering\cite{TSPM,PSTP,SHMamba} and image-text retrieval\cite{MPARN,NSTRN,RCTRN}. Multimodal models can integrate information from diverse modalities and are theoretically expected to outperform their unimodal counterparts. However, recent studies reveal a common issue in multimodal learning: modality bias. During joint training, the advantaged modality often dominates the backpropagation, leading to imbalanced optimization. To address this, OGM-GE\cite{OGM} dynamically tracks each modality’s marginal contribution to the joint objective during training and adaptively regulates the backward gradients to suppress the dominant one, aiming for balanced convergence. DGL \cite{DGL} tackles optimization conflicts by truncating the multimodal loss gradients before they reach the encoders. It then trains the encoders with unimodal losses using a modality-zeroing strategy, thereby decoupling the optimization processes of the encoders and the fusion module. InfoReg\cite{InfoReg} adopts an information-theoretic approach: it leverages the trace of Fisher information\cite{Fisher} to estimate each modality’s information absorption rate during the early critical learning phase and applies an adaptive regularizer to decelerate modalities that have already absorbed sufficient information, thereby enhancing cross-modal optimization.

\begin{figure*}
  \centering
  \subfloat[RLC-base]{\includegraphics[width=0.33\linewidth]{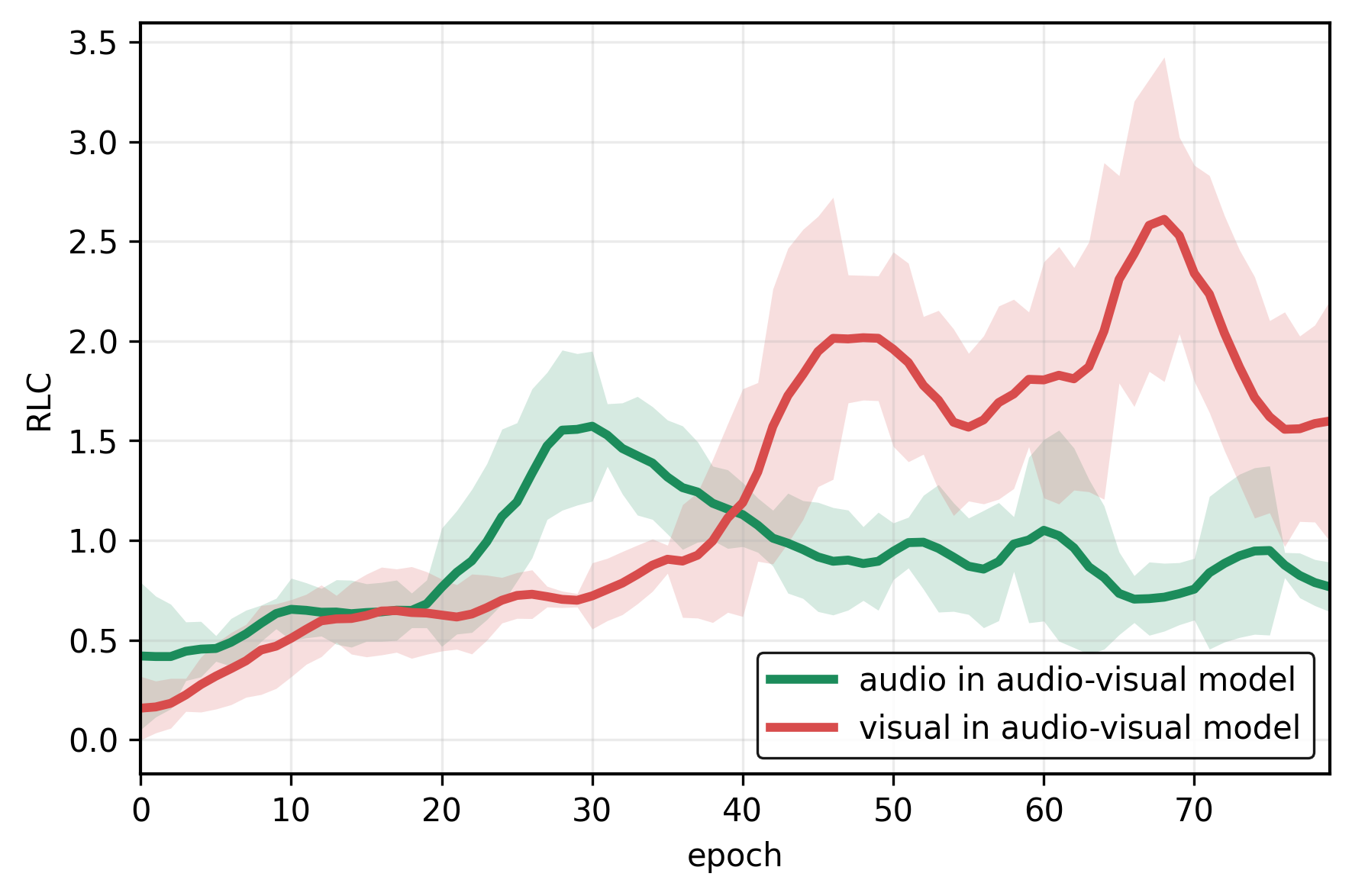}\label{fig:sub1}}
  \hfill % 控制子图之间的空间
  \subfloat[DGR-base]{\includegraphics[width=0.33\linewidth]{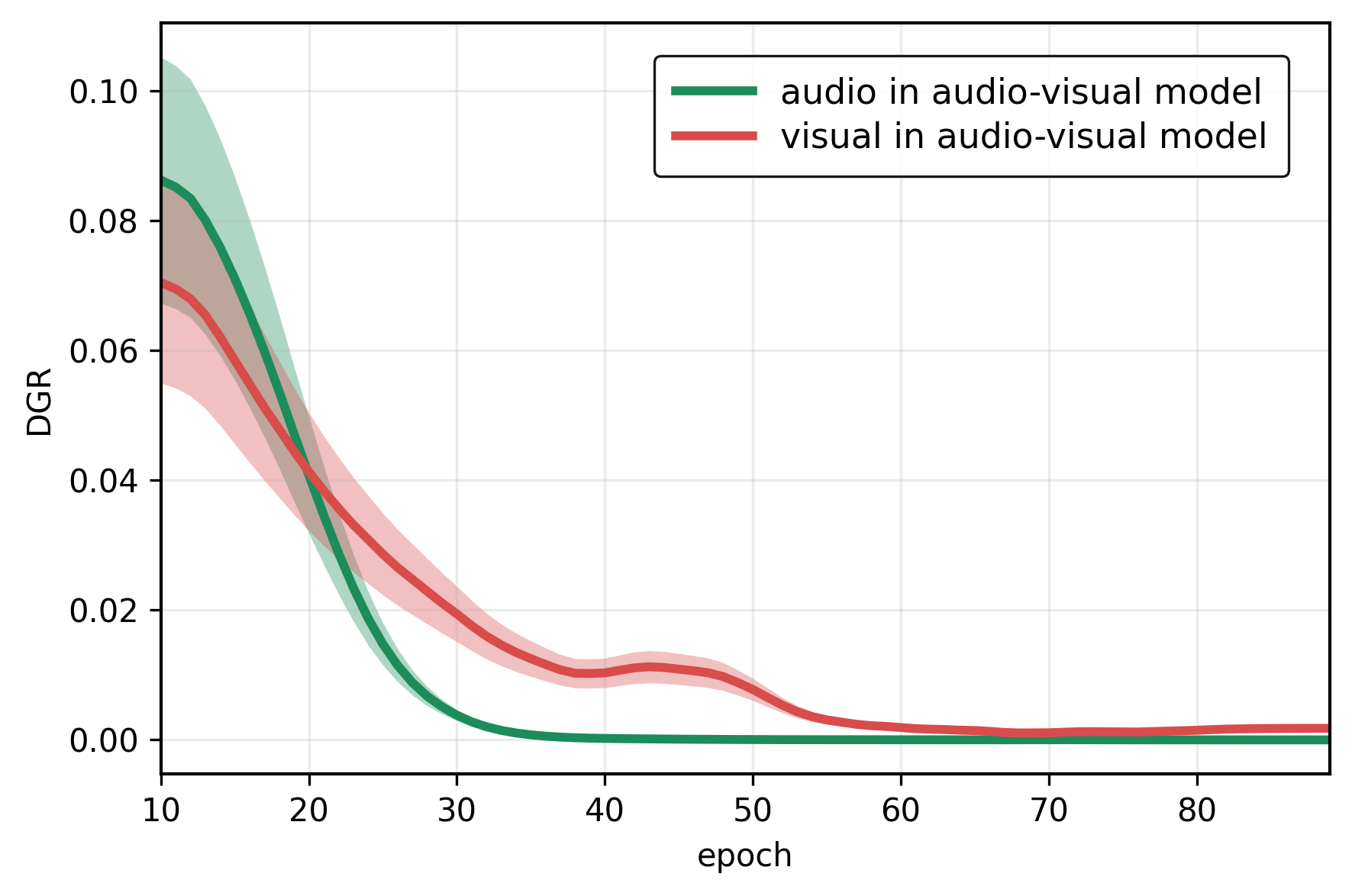}\label{fig:sub2}}
  \subfloat[RS-base]{\includegraphics[width=0.33\linewidth]{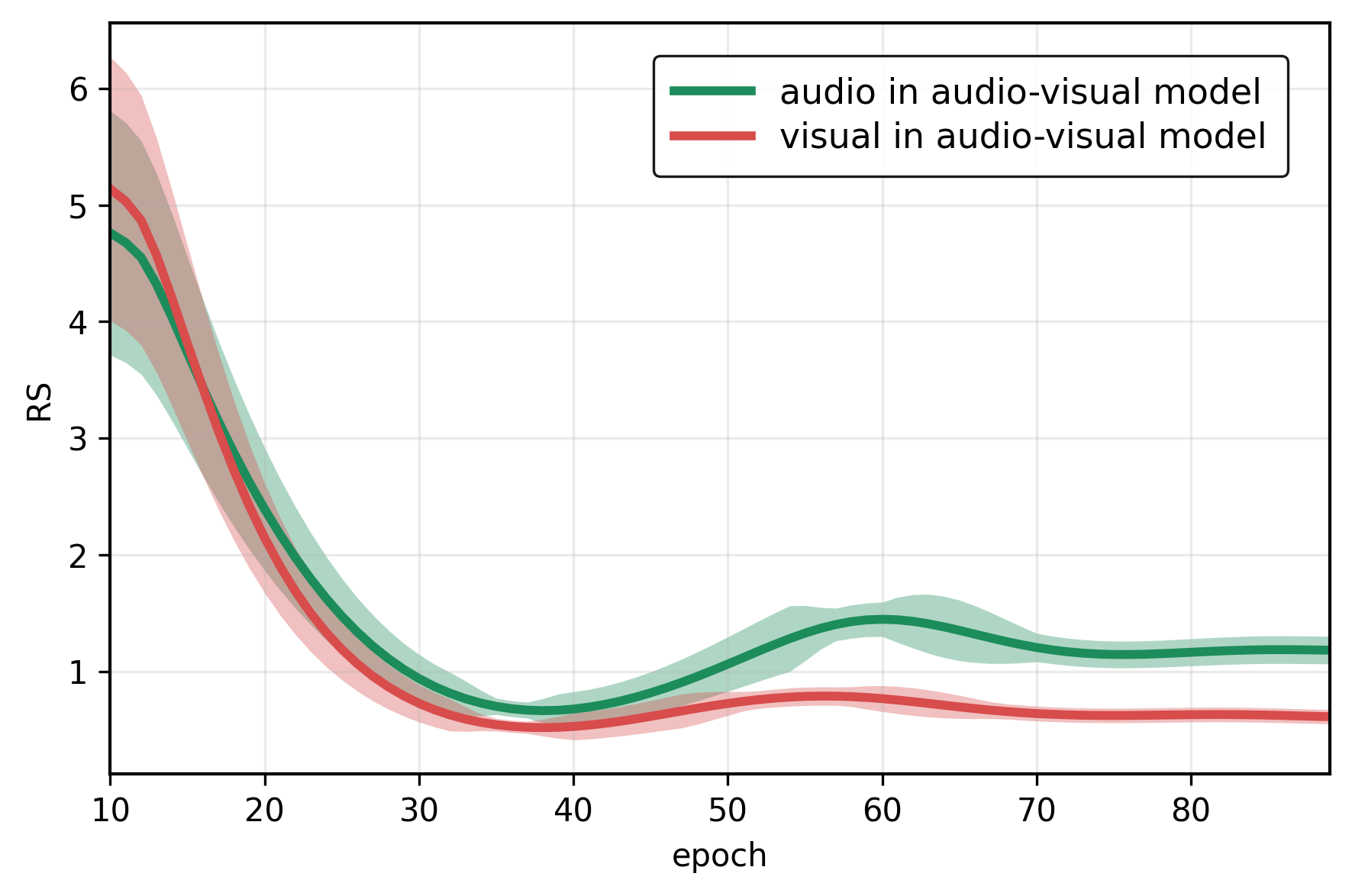}\label{fig:sub3}}
  \caption{Redundancy metrics under the baseline model on CREMA-D: \textbf{(a).} Representation-to-logit coupling (RLC) traces how strongly each modality drives the logits. \textbf{(b).} Effective gain growth rate (DGR) measures the rate of acquiring new, non-overlapping information. \textbf{(c).} Redundancy score (RS) quantifies redundancy.}
  \label{R}
\end{figure*}

Despite notable progress in addressing modality imbalance, several key challenges remain. \textbf{The first challenge is redundancy drift.} In the later stages of training, when the overall loss plateaus, prolonged dominance of one modality during backpropagation can lead to a redundant phase. A typical sign is that changes in the representation result in little or no improvement to the logits. As shown in Fig. \ref{fig:sub1}, we continuously monitor the ratio \(k=\frac{\left \| f_t-f_{t-1} \right \|_2}{\left \| z_t-z_{t-1}  \right \|_2+\varepsilon}\), which measures representation-to-logit coupling. A smaller \(k\) indicates that the representation is changing while the classifier head produces no meaningful response. Early in training, gradients update mainly along task-relevant directions and the classifier head gains effectively, so the advantaged modality (audio) has a higher \(k\). As training continues, the prolonged dominance of the advantaged modality causes updates to drift into low-gain directions of the classifier head. The \(k\) value keeps decreasing and approaches zero over a certain interval, which forms a redundant phase. In contrast, the weaker modality is still learning and therefore maintains a relatively high \(k\). \textbf{The second challenge is the unified gradient suppression that ignores semantics and orientation.} Many existing approaches\cite{InfoReg, OGM, W_Tpami} uniformly downscale the advantaged modality in the joint gradient, suppressing both redundant components and task-critical semantics while failing to distinguish between shared and modality-specific features. When the task at a particular moment depends primarily on either vision or audio, such unified suppression may compromise useful unimodal signals. At the same time, weakening the gradient along the main descent direction can increase inter-gradient angles. This may lead to descent deflection and convergence oscillations, which slow down optimization and degrade the final performance. As shown in Fig. \ref{fig:sub2} and Fig. \ref{fig:sub3}, we monitor the effective gain growth rate and the redundancy score for both modalities. Early in training, as the model quickly learns task features, the space to further increase the correct-class probability per unit time shrinks, so the effective gain growth rate drops rapidly as representations take shape. In the middle to late stages of training, the advantaged modality exhausts its usable semantics and its effective gain growth rate remains close to zero for an extended period. Meanwhile, the weaker modality still contains learnable signals, resulting in a slight rebound in the effective gain growth rate. In the early stage of training, the model has not yet established robust invariances and remains sensitive to small input perturbations, which leads to a high redundancy score. During mid to late training, the redundancy score of the advantaged modality rebounds and stays high, while that of the weaker modality remains low. The co-occurrence of low effective gain growth rate and high redundancy score during the same period defines what we refer to as the redundant phase.

To address these challenges, we propose an online adaptive regulation strategy based on the information bottleneck principle \cite{15IB,17IB}, targeting late-stage imbalance control and redundancy suppression. Specifically, we construct a redundant phase monitor by combining the effective gain growth rate with redundancy. We trigger intervention only when the dominant modality exhibits high redundancy and effective semantic gain has stalled, which avoids unnecessary regularization overhead. Second, we introduce an information-gating mechanism to assess the current reliance on cross-modal semantics. When the prediction primarily depends on key semantics from a single modality, the gate automatically deactivates the suppression term to preserve modality-specific information. Finally, we apply rate-limited suppression to the dominant modality within the orthogonal complement of the subspace defined by the joint multimodal gradients. We first project its gradient onto this orthogonal complement, then scale it based on the measured redundancy, thereby attenuating only task-irrelevant drift components. Notably, our strategy requires no additional supervision or specialized architectures and can be seamlessly integrated with existing methods. Experimental results demonstrate that the proposed method effectively suppresses modality redundancy and substantially alleviates optimization imbalance.

Our main contributions are summarized as follows:
\begin{itemize}
\item We identify and characterize a phenomenon we term redundancy drift, which emerges in the later stages of training. We use redundant phase detection to quantify and localize this drift and use this phase as a trigger for subsequent intervention.

\item We propose adaptive redundancy regulation mechanism. We design an information gating mechanism to identify cross-modal semantic dependencies. Then, within the orthogonal complement of the joint gradient subspace, we apply speed-limiting suppression to the dominant modality, selectively suppressing task-irrelevant redundant drift.
\item Experimental results show that our method consistently improves the accuracy on most datasets and significantly alleviates modality bias. Ablation studies further validate the effectiveness of each component.
\end{itemize}

In Section \ref{A}, we discuss the related work. We introduce multimodal learning, imbalanced multimodal learning, and the information bottleneck in deep learning. Section \ref{B} introduces our proposed RedReg method and related algorithms. Section \ref{C} demonstrates the effectiveness of our method through experimental results and visualizations. Section \ref{E} discusses the advantages of this approach and future research directions. Finally, Section \ref{D} summarizes our research results, highlighting the main findings and contributions.

\section{Related Work}
\label{A}
\subsection{Multimodal Learning} 
Multimodal learning aims to align heterogeneous signals such as images, speech, and text into a shared semantic space, enabling complementary and synergistic use of information. One line of existing work focuses on enhancing classical unimodal tasks by leveraging cross-modal cues, thereby improving the robustness and generalization of tasks such as action recognition\cite{song1,sun1,li1}, image retrieval\cite{song2,sun2,yang1,MPARN,NSTRN,RCTRN} and video representation\cite{SpiVG,sun3,li2,li3,yuan3,HcPCR,liu1,liu2,liu3}. Another line of research proposes new problem formulations and evaluation frameworks centered on cross-modal interaction, such as audio-visual learning\cite{TSPM,PSTP,STFT,MDST++,MDFT,MSTR,SHMamba}, text-to-image\cite{zhou1,yuan1,yang2} and text-to-video generation\cite{yang3,song3,RMARN}. These are designed to assess model alignment and reasoning under more complex spatiotemporal and semantic constraints. However, these studies primarily focus on architectures and fusion mechanisms, emphasizing representation alignment and information exchange. Most of them rely on static fusion or globally fixed modulation throughout training, which makes it difficult to identify and address the ineffective complexities that emerge during the later stages of training. 

To address this issue, we propose an online and adaptive modulation strategy based on the information bottleneck principle, which aims to mitigate late-stage training imbalance and reduce redundancy. 

\subsection{Imbalanced Multi-Modal Learning} 
In principle, multimodal models should outperform their unimodal counterparts by leveraging data from multiple perspectives. However, recent research\cite{OGM, W_Tpami, InfoReg,UCD,MDPFL,DGL,AMSS++,G-Blending,DRBML,ACA} suggests that inter-modal learning imbalance hinders the effective training of multimodal networks, which has led to increased interest in understanding its causes. Recent studies address this imbalance from multiple perspectives. Some identify the key issue as early-stage disparities in information intake and propose adaptive suppression of the dominant modality's acquisition rate within a critical learning window, as seen in InfoReg\cite{InfoReg}. Others focus on online gradient-level modulation. For instance, OGM-GE\cite{OGM} rescales the dominant modality’s gradients and injects noise to restore generalization, while DGL\cite{DGL} decouples the gradient paths of unimodal and multimodal losses to resolve conflicts between encoders and fusion modules. At the subnetwork level, AMSS++\cite{AMSS++} leverages modality salience and Fisher information–based non-uniform sampling to selectively update only the foreground subnetwork during backpropagation, thereby allocating more capacity to the weaker modality. Other approaches target classifier capacity. Techniques such as sustained boosting and adaptive classifier allocation aim to continuously strengthen the classifier associated with the weak modality, as seen in Sustained Boosting and ACA\cite{ACA}. In addition, G-Blending\cite{G-Blending} adjusts modality weights during training based on generalization quality, using the overfitting-to-generalization ratio as a guiding signal. Overall, these methods mitigate the modality imbalance effect in various dimensions, including information acquisition, gradients, parameters, and classifiers. They have demonstrated consistent improvements in audio-video, vision-language, and trimodal tasks.

In contrast to previous approaches, we propose an online and adaptive regulation strategy based on the information bottleneck principle, focusing on mitigating late-stage imbalance and suppressing redundancy. Our method avoids introducing additional complex modules. Instead, it utilizes observable learning signals to dynamically regulate the flow of redundant information, suppress overfitting in the dominant modality, and create optimization space for the weaker modality. This leads to more robust multimodal balance.

\begin{figure*}
    \centering    \includegraphics[width=1.0\linewidth]{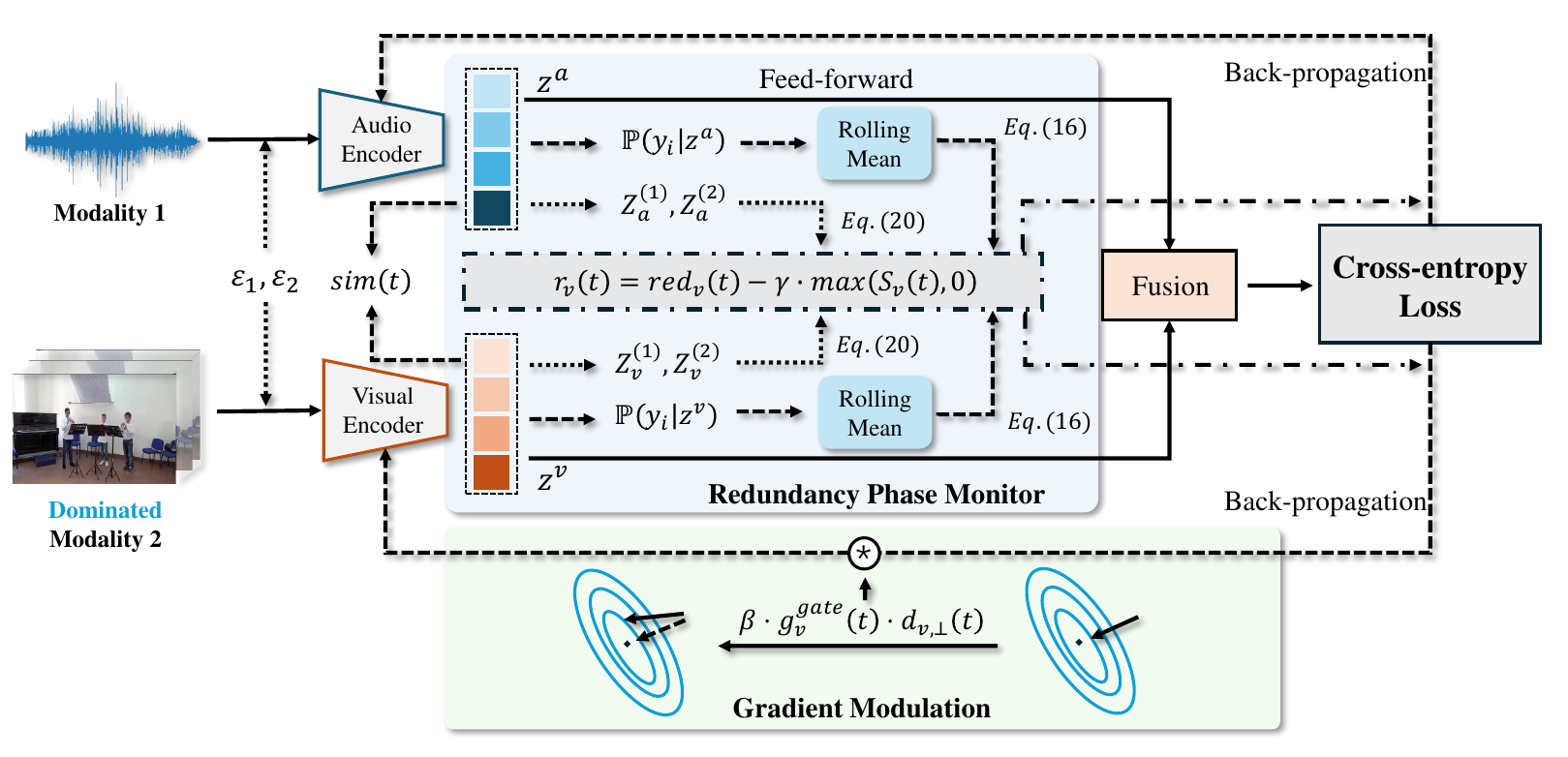}
    \caption{Overview of RedReg. The figure illustrates the architecture and training strategy of RedReg. During the feed-forward state, RedReg first determines the dominant modality, then evaluates whether a redundant state occurs based on the modality growth rate and the degree of redundancy, and finally applies redundancy regulation during the back-propagation pass.} 
    \label{fig2}
\end{figure*}
\subsection{The Information Bottleneck in Deep Learning} 
The Information Bottleneck (IB) framework aims to learn a minimal sufficient representation. It does so by compressing the mutual information between the input and the representation, $I(X; Z)$, while preserving the task-relevant mutual information $I(Z; Y)$. This reduces redundancy and enhances generalization, forming the theoretical foundation of our method. The classic IB framework was introduced by Tishby et al. \cite{IB1}, who proposed a variational principle and a solution method inspired by rate-distortion theory. VIB\cite{VIB} extended this idea into the end-to-end trainable Variational Information Bottleneck, incorporating learnable noise injection and a KL regularizer as a tractable upper bound on $I(X; Z)$. Achille and Soatto \cite{IB3} connected the IB principle with regularization and variational autoencoders (VAEs) through the lens of information dropout, showing how noise injection promotes invariance and generalization. DIB\cite{DIB} proposed the Deterministic Information Bottleneck, which replaces mutual information with entropy to better capture the notion of compression. They also derived analytical properties that were later extended to a geometric clustering perspective. In recent multimodal studies, the Multimodal Information Bottleneck (MIB) \cite{MIB} aims to minimize redundant information about the original multimodal inputs in the fused representation while preserving task-relevant signals. This approach has shown improvements in robustness in applications such as multi-sensor reinforcement learning and representation compression. 

In contrast to these approaches, which primarily focus on static representation learning or global regularization, our method applies the IB principle to training dynamics for online control. To address late-stage imbalance and redundancy amplification, we leverage observable learning signals to dynamically regulate the flow of redundant information. By curbing overfitting in the dominant modality, we create extra room for optimization in the weaker ones and achieve robust multimodal balance without adding complex modules.

\section{Methodology}
\label{B}
\subsection{Problem Introduction} 
This work studies multimodal classification trained end to end using only a joint multimodal cross-entropy loss. A common phenomenon in multimodal tasks is that a dominant modality suppresses the weaker one, so the fused model can even underperform certain unimodal methods (see \cite{OGM}). We argue that the primary root cause is that, in late training, the dominated modality encounters an information bottleneck and accumulates redundancy: its effective gain with respect to the labels stops increasing, yet it continues to complicate the representation along directions that are first-order ineffective for the task, which heightens sensitivity to the input and crowds out the weaker modality’s learning channel.

Given a dataset \(D=\{\left(x_n,y_n\right)\}_{n=1}^N\) , each sample comprises $M$ modalities $x_n=\left(x_n^1,x_n^2,\ldots,x_n^M\right)$, with labels $y_n\in\left\{1,\cdots,K\right\}$ for $K$ classes. For each modality $m\in{1,\cdots,M}$, an encoder $\phi_m(w_m,\cdot)$ maps inputs to a representation space:
\begin{equation}
   z_n^m=\phi_m(w_m,x_n^m),
\end{equation}
where $\phi_{m}$ is the encoder for modality $m$ with parameters $w_{m}$ and $x_{n}^{m}$ is the $m$-th modality input of sample $n$. $z_{n}^{m} \in \mathbb{R}^{d_{m}}$ is its representation of dimension $d_{m}$. The modalities are then fused. Here we use simple concatenation as an example:
\begin{equation}
   z_n=concat\left(z_n^1,\ldots,z_n^M\right),
\end{equation}
where $concat(\cdot)$ stacks the per-modality representations along the feature dimension to form the fused vector $z_n \in \mathbb{R}^d$ with $d = \sum_{m=1}^M d_m$. The fused representation is fed to a linear classification head:
\begin{equation}
   f(z_n)=W\cdot\ z_n+b,
\end{equation}
where $f(z_n)\in\mathbb{R}^K$ are the class logits. $W\in\mathbb{R}^{K\times d}\mathrm{~and~}b\in\mathbb{R}^K$ are the classifier weight and bias. $K$ is the number of classes. The product $W\cdot z_n$ is a matrix-vector multiplication. The model is optimized with the joint multimodal cross-entropy loss:
\begin{equation}
\begin{aligned}
\mathcal{L}_{\text{joint}}(w,\theta) 
  &= \frac{1}{N}\sum_{n=1}^{N}\!\left(-y_n^{\top}\log\!\bigl[\operatorname{softmax}(f(z_n))\bigr]\right) \\
  &= \frac{1}{N}\sum_{n=1}^{N}\!\left(-y_n^{\top}\log p_n\right),
\end{aligned}
\end{equation}
where $N$ is the number of training samples. $w=\{w_1,\dots,w_M\}$ collects all encoder parameters. $\theta=\{W,b\}$ are the classifier parameters. $p_n=\text{softmax}(f(z_n))\in\mathbb{R}^K$ is the vector of class probabilities. $y_n\in\{0,1\}^K$ is the one-hot label. $\log(\cdot)$ is the natural logarithm applied element-wise.

Without loss of generality, we take audio to be the dominated modality. Let $t$ denote the time step and $B$ the batch size. By the chain rule, the encoder gradients at time $t$ can be written as:
% \begin{equation}
% \begin{aligned}
% g_a(t) &= \nabla_{w_a}\,\mathcal{L}_{\text{joint}}(w_a,\theta_a)\\
%        &= \frac{1}{N}\sum_{n=1}^{N} J_a(x_n^a)^\top W_a^\top (p_n - y_n).
% \end{aligned}
% \end{equation}

% \begin{equation}
% \begin{aligned}
% g_v(t) &= \nabla_{w_v}\,\mathcal{L}_{\text{joint}}(w_v,\theta_v)\\
%        &= \frac{1}{N}\sum_{n=1}^{N} J_v(x_n^v)^\top W_v^\top (p_n - y_n).
% \end{aligned}
% \end{equation}

\begin{equation}
\begin{aligned}
g_a(t) &= \nabla_{w_a}\,\mathcal{L}_{\text{joint}}(w_a,\theta_a)\\
       &= \frac{1}{N}\sum_{n=1}^{N} J_a(x_n^a)^\top W_a^\top (p_n - y_n),
\end{aligned}
\end{equation}

% \begin{equation}
% {g_a\left(t\right)=\nabla}_{w_a}\mathcal{L}_{joint}\left(w_a,\theta_a\right)=\frac{1}{N}\sum_{n=1}^{N}{{J_a\left(x_n^a\right)}^TW_a^T{{(p}_n-y_n)}},
% \end{equation}

% \begin{equation}{g_v\left(t\right)=\nabla}_{w_v}\mathcal{L}_{joint}\left(w_v,\theta_v\right)=\frac{1}{N}\sum_{n=1}^{N}{{J_v\left(x_n^v\right)}^TW_v^T{{(p}_n-y_n)}},
% \end{equation}

\begin{equation}
\begin{aligned}
g_v(t) &= \nabla_{w_v}\,\mathcal{L}_{\text{joint}}(w_v,\theta_v)\\
       &= \frac{1}{N}\sum_{n=1}^{N} J_a(x_n^v)^\top W_v^\top (p_n - y_n),
\end{aligned}
\end{equation}
where $J_m = \partial z_n^m / \partial w_m$ and $W_m$ denotes the corresponding block of the linear head associated with modality $m$.

In the early stage of training, the advantaged modality $m_a$ dominates optimization because of its stronger correlation with the labels \cite{InfoReg}. Its gradient is much larger than that of the weak modality, which causes its parameters and the classifier block weights $W_a$ to grow faster:

\begin{equation}
\lVert g_a(t)\rVert \gg \lVert g_v(t)\rVert \quad
\text{and}\quad
\frac{\lVert W_a(t+\Delta t)\rVert}{\lVert W_v(t+\Delta t)\rVert} \uparrow,
\end{equation}
where $g_{a}(t)$ and $g_{v}(t)$ are the gradients of the loss with respect to encoder parameters $w_{a}$ and $w_{v}$ at training step $t$. $\left\|\cdot\right\|$ is the Euclidean norm. $W_{a}$ and $W_{v}$ are the classifier blocks that act on the representations of modalities $a$ and $v$. 

The dominated modality quickly reduces the loss and gains a higher fusion weight, thereby taking the lead and forming positive feedback that further dilutes the weak modality’s gradient pathway. When training reaches the later stage, the dominated modality has already pushed the loss to a very low level, so the classification residual 
$e_n = p_n - y_n \approx 0$:
\begin{equation}
\left \| g_a (t) \right \| =\left\| \frac{1}{N}\sum_{n=1}^{N} J_a(x_n^a)^\top W_a^\top e_n \right\| \approx 0.
\end{equation}

At this point, first-order descent directions are nearly exhausted. Nevertheless, perturbations that are not directly driven by the gradient in the optimization process (for example, the noise of small step SGD) can still push $w_a$ to make ineffective but nonzero micro updates. Let a small update $\delta w_a$ induce the following change in the representation space:

\begin{equation}
\Delta z_n^a\approx J_a(x_n^a)^T \delta w_a,
\end{equation}
where $\delta w_a$ is a small change of the encoder parameters of modality $a$. $\Delta z_n^a$ is the resulting first-order change in its representation. We focus on a class of changes that keep the logits unchanged. There exist $\Delta z_n^a$ that lie in the zero response subspace of the classifier head for the dominated modality:

\begin{equation}
\mathcal{N}_a:=\left \{ u\in\mathbb{R} ^{dimz^a}:W_au=0 \right \} ,
\end{equation}
where $\mathcal{N}_a$ is the null space of $W_a\mathrm{~and}\dim z^a$ is the dimensionality of $z^a$. If $\Delta z_n^a \in \mathcal{N}_a$, then to first order $\triangle f_n$:
\begin{equation}
\triangle f_n=W_a\triangle z_n^a\approx 0\Rightarrow \triangle \mathcal{L}_{joint}\approx 0,
\end{equation}
which means the logits and the loss are almost unchanged. Such updates do not improve discrimination but cause lateral drift in the representation space. In other words, the model is in an information bottleneck, as shown in Fig. \ref{R}.

\subsection{Information Bottleneck in Multimodal Learning} 
Inspired by the information bottleneck principle\cite{15IB,17IB,aaaiIB}, we investigate its utility in the training of multimodal models. To address the phenomenon that the dominated modality reaches an information bottleneck in multimodal learning, we regulate the effective gain and redundancy of the information rich modality throughout training. See the Algorithm \ref{alg} for details. When modality information becomes redundant, we adjust the rate limiting direction and, through co-information gating mechanism, slow its information acquisition so as to avoid compressing subsequent useful semantics. As shown in Fig. \ref{fig2}, our method has the following core components: Redundant Phase Monitor and Co-info Gating.
\subsubsection{Redundant Phase Monitor}
The information bottleneck aims to learn a minimal sufficient intermediate representation $Z$. It seeks to retain task relevant information $I(Z; Y)$ while discarding input irrelevant redundancy $I(Z; X)$. Its classical Lagrangian form is:

\begin{equation}
\min_{p(z|x)}\quad\mathcal{L}_{\mathrm{IB}}=I(Z;X)-\beta I(Z;Y),
\end{equation}
where $I(\cdot\,;\,\cdot)$ denotes mutual information. A smaller loss means $Z$ preserves little information about $X$ while still retaining enough information to predict $Y$. The trade off between the two mutual information terms is controlled by the Lagrange multiplier $\beta$.

Let the multimodal input be $ X = (X^a, X^v)$. Inspired by the information bottleneck, we aim to learn a task sufficient and compact multimodal representation $ Z = [Z^a : Z^v]$ that retains label relevant information $I(Z; Y)$ while discarding input irrelevant redundancy $I(Z; X)$. The multimodal IB objective can be written as:
\begin{equation}
\min\quad\alpha_a\cdot I(Z^a;X^a)+\alpha_v\cdot I(Z^v;X^v)-\alpha\cdot I([Z^a;Z^v];Y).
\end{equation}

However, directly estimating mutual information on high-dimensional deep representations is difficult. First, practical encoders are typically deterministic continuous mappings $Z = f_w(X)$, which gives $p(z \mid x) = \delta\!\left(z - f_w(x)\right)$ and $H(Z \mid X) = -\infty$. Thus $I(Z; X) = H(Z) - H(Z \mid X)$ often diverges or has no upper bound in the continuous case. Second, in the multimodal setting we must estimate $I(Z^a; X^a)$, $I(Z^v; X^v)$, and $I([Z^a : Z^v]; Y)$, while handling cross-modal coupling and matching complexity. Common contrastive estimators require $O(N^2)$ cost. This makes direct optimization unstable and not scalable. Therefore we use observable and low-cost IB proxy quantities for online decision making. To distinguish effective learning from redundant accumulation, we measure two observable quantities: the effective gain growth rate $s_m (t)$ (an observable proxy
for $I(Z_m;Y)$) and the redundancy score $red_m(t)$ (a lightweight proxy for $I(Z_m;X)$).
\begin{algorithm}[t]
\caption{Training Pipeline of RedReg}
\label{alg}
\begin{algorithmic}[1] % Enables line numbering
\fontsize{9}{10}\selectfont
\Require Training dataset \(D\), number of epochs \(T\), number of batches \(B\) of each epoch, sliding length \(L\), hyperparameter \(\gamma\), $\beta$, $R$, \(\tau_{\min}\), \(\tau_{\max}\).
\Ensure Trained model parameters.
\For{\(t = 0,1,\ldots,T-1\)}
  \If{\(t < 2\)}
    \State Update model parameters;
    \State \textbf{continue};
  \EndIf
  \State \hfill \textcolor{gray}{/*** Determination of redundant phase  ***/}
  \State Calculate \(p_m\) by Eq.~\eqref{p_m};
  \State Calculate \(s_m(t)\) by Eq.~\eqref{smt};
  \State Calculate \(red_m(t)\) by Eq.~\eqref{redm};
  \State Calculate \(r_m(t)\) by Eq.~\eqref{r_m(t)};

  \State \hfill \textcolor{gray}{/*** Gating and direction protection ***/}
  \State Calculate \(\mathrm{sim}(t)\) by Eq.~\eqref{sim(t)};
  \State Calculate \(\tau(t)\) by Eq.~\eqref{tau(t)};
  \State Obtain \(g_m^{\mathrm{gate}}(t)\) by Eq.~\eqref{gate};
  \If{\(g_m^{\mathrm{gate}}(t) > 0\)}
    \State Calculate \(d_{m,\perp}(t)\) by Eq.~\eqref{d(t)};
    \State Update model parameters by Eq.~\eqref{update};
  \EndIf
  \State Update model parameters;
\EndFor
\end{algorithmic}
\end{algorithm}

For modality $m \in \{a, v\}$, define at training step $t$ the branch correct class probability as:

\begin{equation}
p_m=\frac{1}{B}\sum_{i=1}^{B}{\mathbb{P}\left(y_i\middle| z_i^m\left(t\right)\right)},
\label{p_m}
\end{equation}
where $B$ is the mini-batch size. $\mathbb{P}(y_{i} \mid z_{i}^{m}(t))$ is the model probability assigned to the ground-truth class $y_{i}$ given the branch representation $z_{i}^{m}(t)$ from modality $m$ at step $t$. $z_{i}^{m}(t)$ is computed by the encoder of modality $m$ followed by the branch classifier.
The smoothed mean is defined based on a sliding window of length $L$ :
\begin{equation}
{\bar{p_m}}^{\left(t\right)}=\frac{1}{L}\sum_{k=t-L+1}^{t}{p_m\left(k\right)}.
\end{equation}

To characterize the speed of learning, we construct a dimensionless growth rate:
\begin{equation}
S_m\left(t\right)=\frac{{\bar{p_m}}^{\left(t\right)}-{\bar{p_m}}^{\left(t-1\right)}}{|{\bar{p_m}}^{\left(t-1\right)}|+\varepsilon}.
\label{smt}
\end{equation}
where $\varepsilon > 0$ is a small constant for numerical stability. $S_m(t) > 0$ indicates that the model is still learning label relevant effective information, while $S_m(t) \approx 0$ indicates that information gain has stalled.

To obtain a redundancy proxy, we apply two weak augmentations with small noise to the same input:

\begin{equation}
\begin{aligned}
x^{(1)} &= x + \varepsilon_1, x^{(2)} = x + \varepsilon_2,
\\
\varepsilon_i \sim &\mathcal{N}(0,\sigma^2 I),\quad i\in\{1,2\},
\end{aligned}
\end{equation}
and pass them through the encoder to get two views:

\begin{equation}
Z_m^{(1)}=\phi_m(w_m,x^{(1)}),
\end{equation}
\begin{equation}
Z_m^{(2)}=\phi_m(w_m,x^{(2)}),
\end{equation}
where $x$ is an input of modality $m$. $\varepsilon_{1}$ and $\varepsilon_{2}$ are independent Gaussian noises with variance $\sigma^{2}$ and identity covariance. $\phi_{m}(w_{m},\cdot)$ is the encoder of modality $m$ with parameters $w_{m}$. $Z_{m}^{(1)}$ and $Z_{m}^{(2)}$ are the corresponding representations. We define the normalized consistency error as the redundancy score:

\begin{equation}
red_m(t)=\frac{\left\|Z_m^{(1)}-Z_m^{(2)}\right\|_2^2}{\left\|x^{(1)}-x^{(2)}\right\|_2^2+\varepsilon}.
\label{redm}
\end{equation}

When the noise is sufficiently small and $\phi_m$ is locally linearizable, $\mathbb{E}[\mathrm{red}_m] \propto \|J_m\|_F^2$ with $J_m=\partial z_m/\partial x_m$. In other words, the more sensitive the representation is to the input, the higher the redundancy. 

Combining Eq.(\ref{smt}) and Eq.(\ref{redm}), we can obtain the Redundant Phase Monitor:

\begin{equation}
r_m(t)=red_m(t)-\gamma\max(S_m(t),0),
\label{r_m(t)}
\end{equation}
where $\gamma$ is a non–negative trade–off coefficient that balances the contribution of redundancy and the penalty from sparsity. A larger $\gamma$ enforces stronger suppression on the redundant phase when the sparsity indicator $S_m(t)$ becomes positive, while a smaller $\gamma$ emphasizes the redundancy term itself.

\subsubsection{Co-info Gating}
Although we have defined the redundant phase monitor $r_m(t)$ to determine when the model enters information redundancy and which modality is involved, directly compressing the representation of the dominated modality at that moment (that is, reducing $I(Z_m; X)$) carries a significant risk. If the current sample mainly relies on private semantics, namely discriminative cues visible only in a single modality, indiscriminate compression would damage this useful information and would reduce $I(Z_m; Y)$. To avoid this side effect, we propose a controlled update strategy that gates first and then applies rate limiting. We first use co-information gating mechanism to verify that the cross-modal semantics are truly shared. Only then do we apply rate limiting to the dominated modality within the subspace that is orthogonal to the task gradient. In this way we suppress redundancy while preserving the main descent direction of the task.

Intuitively, it is safer to constrain the dominated modality when the two modalities see the same or similar semantics for the same sample in the current batch. If their points of focus differ greatly and the shared content is low, there is a high probability of private semantics, in which case the dominated modality should be allowed to keep learning. At time step $t$, extract the intermediate features from the two branches (the global vectors before the classifier head):

\begin{equation}
z^a_i(t)\in\mathbb{R}^{d_a},\quad z^v_i(t)\in\mathbb{R}^{d_v},\quad i=1,\ldots,B,
\end{equation}
where $z_i^a(t)$ and $z_i^v(t)$ are the audio and visual global features of sample $i$. $d_a$ and $d_v$ are their feature dimensions. $B$ is the mini-batch size. After $\ell_2$ normalization we obtain $\widehat{z^a_i}(t)$ and $\widehat{z^v_i}(t)$. Using the diagonal cosine similarity of paired samples to measure cross-modal sharing, define the batch level co-information proxy:

\begin{equation}
\mathrm{sim}(t)\;=\;\frac{1}{B}\sum_{i=1}^{B}\,\langle\widehat{z^a_i}(t),\,\widehat{z^v_i}(t)\rangle.
\label{sim(t)}
\end{equation}
where$\langle u,v\rangle=u^\top v$ is the dot product. The cosine form follows from unit-norm features. Since the two branches are not yet aligned early in training, an overly high threshold would block too aggressively. We therefore use an increasing threshold schedule:
\begin{equation}
\begin{aligned}
\tau\left(t\right)&=\tau_{min}+\frac{t}{T}({\tau_{max}-\tau}_{min}),\\
&0<\tau_{min}<\tau_{max}<1,    
\end{aligned}
\label{tau(t)}
\end{equation}
where $\tau_{min}$ and $\tau_{max}$ are hyperparameters. 

To align with redundancy detection, we intervene only when the modality is strong and currently in the redundancy stage. The gating coefficient is:

\begin{equation}
\begin{aligned}
g_m^{\mathrm{gate}}(t)\;=&\;\mathbf{1}\{m=m_{dominated}\}\cdot \mathbf{1}\{r_m(t)> R\}\\
&\cdot\mathbf{1}\{\mathrm{sim}(t)\ge\tau(t)\},    
\end{aligned}
\label{gate}
\end{equation}
where $R$ is a predefined threshold parameter that controls the minimum redundancy indicator required for gate activation, preventing trivial or noisy activations.

Even when the gate is open, directly adding a rate limiting direction to the task gradient may still change the first order descent of the main task. Therefore, we require the intervention term to always lie in the orthogonal complement of the task gradient so that the descent component most relevant to the task is unchanged. Let the gradient of the joint loss with respect to the branch parameters be:
\begin{equation}
g_m(t)=\nabla_{w_m}\mathcal{L}_{\text{joint}}(t).
\end{equation}

We use an anchor of the dominated modality parameters $w_m^{(g)}(t)$ to define a slowly varying direction:

\begin{equation}
d_m(t)=w_m(t)-w_m^{(g)}(t),
\end{equation}
where $w_m^{(g)}(t)$ is a slowly updated anchor.
Then we project this direction $d_m$ onto the orthogonal complement of $g_m$:
\begin{equation}
P_{\perp}(g_m) = I - \frac{g_m g_m^{\top}}{\lVert g_m\rVert_2^{2} + \varepsilon},
\end{equation}
\begin{equation}
d_{m,\perp}(t) = P_{\perp}(g_m)\, d_m(t),
\label{d(t)}
\end{equation}
where $I$ is the identity matrix. 

Combining the gating coefficients in Eq. (\ref{gate}) gets the controlled update vector:
\begin{equation}
\widetilde{g}_m(t)=g_m(t)+\beta\cdot g_m^{\text{gate}}(t)\,d_{m,\perp}(t),
\label{update}
\end{equation}
where $\beta$ is the gradient sensitivity coefficient. A higher value of $\beta$ strengthens the regularization, preventing the dominant mode from over-dominantly dominating the training signal. Note that since $\left \langle g_m ,d_{m,\perp} \right \rangle =0$, the rate limiting term does not change the first order descent component of the main task. It only applies braking along directions orthogonal to the task, which maximally avoids harming $I(Z;Y)$:
\begin{equation}
\langle\tilde{g}_m(t),g_m(t)\rangle=\|g_m(t)\|_2^2\geq0.
\end{equation}

\begin{figure*}
  \centering
  \subfloat[RLC-RedReg]{\includegraphics[width=0.33\linewidth]{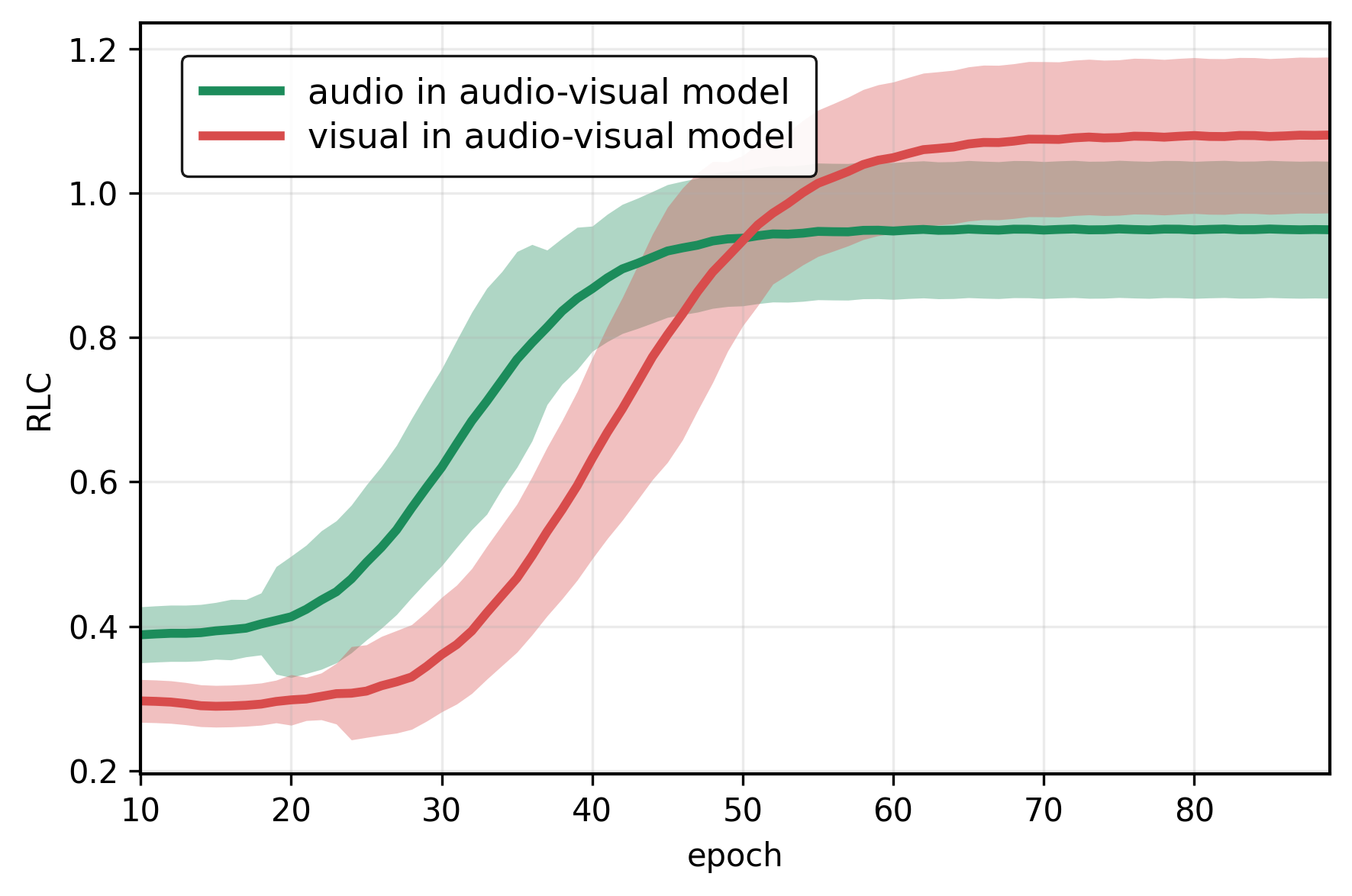}\label{e(a)}}
  \hfill % 控制子图之间的空间
  \subfloat[DGR-RedReg]{\includegraphics[width=0.33\linewidth]{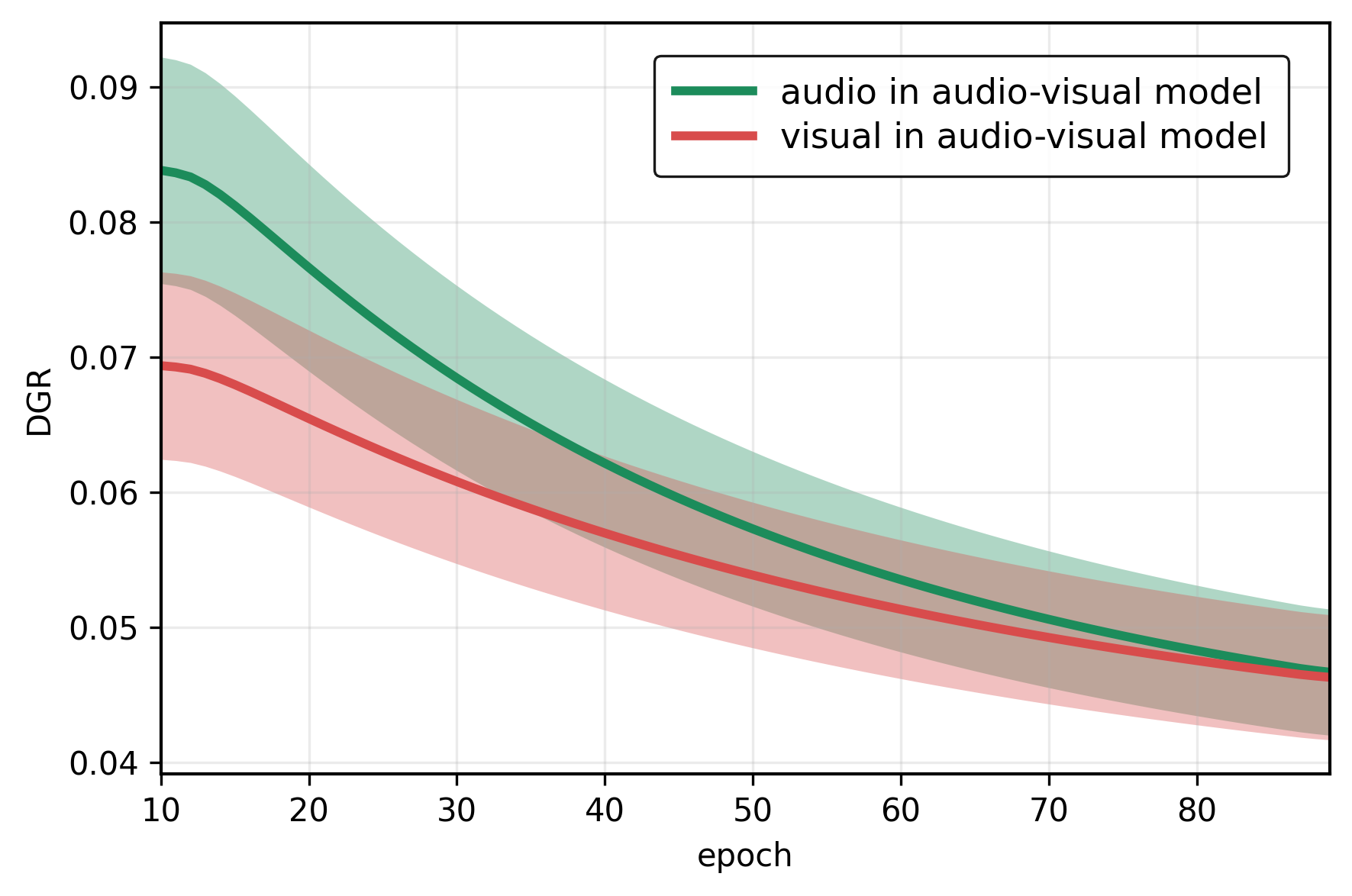}\label{e(b)}}
  \subfloat[RS-RedReg]{\includegraphics[width=0.33\linewidth]{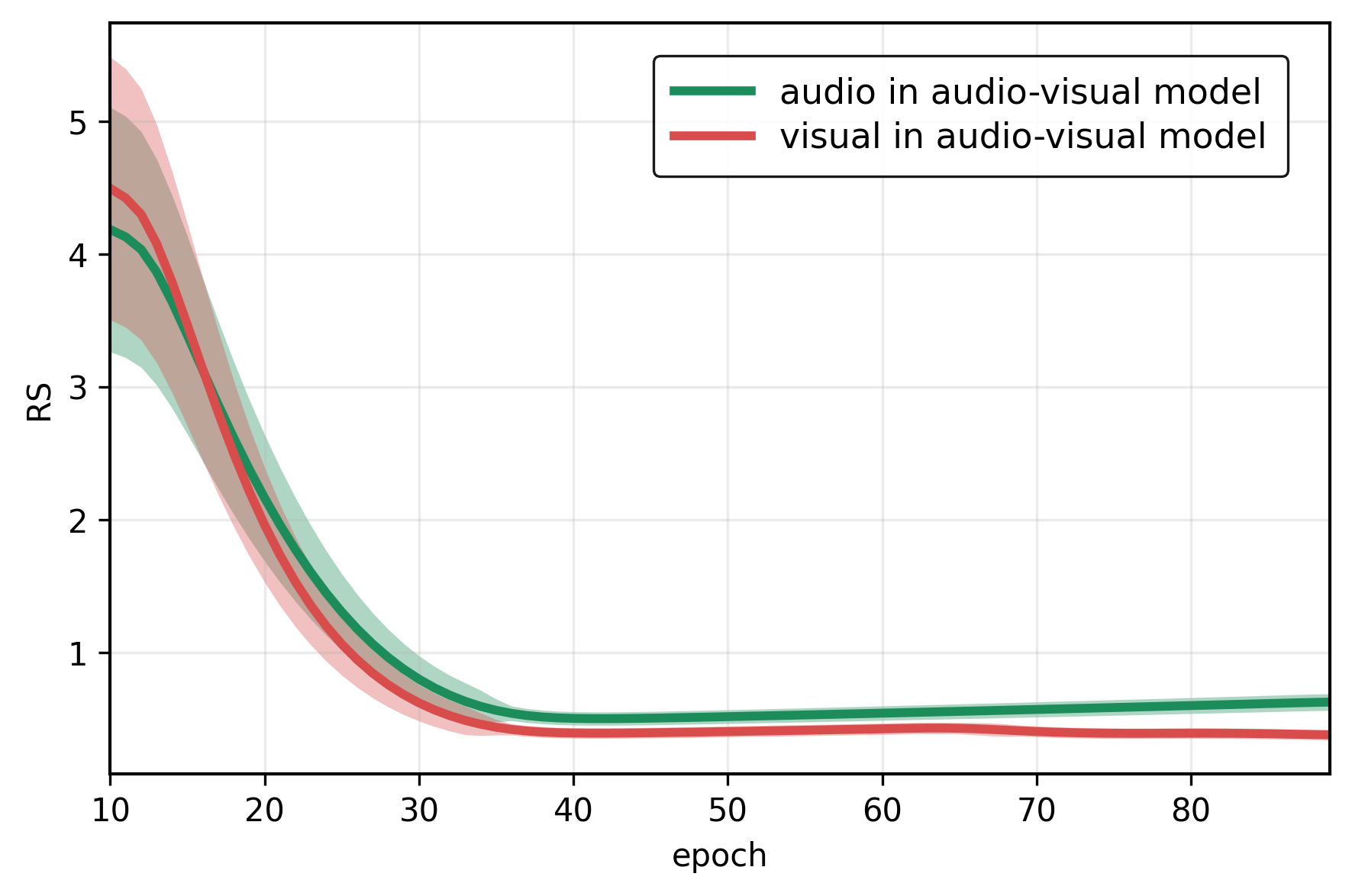}\label{e(c)}}
  \caption{Redundancy metrics under RedReg model on CREMA-D: \textbf{(a).} Representation-to-logit coupling (RLC) traces how strongly each modality drives the logits. \textbf{(b).} Effective gain growth rate (DGR) measures the rate of acquiring new, non-overlapping information. \textbf{(c).} Redundancy score (RS) quantifies redundancy.}
  \label{ER}
\end{figure*}

\section{Experiments}
\label{C}

\subsection{Datasets Statistics}
\textbf{CREMA-D}\cite{CREMA-D} is an audio-visual dataset for emotion recognition in acted speech and facial expressions, containing 7,442 clips from 91 actors across six emotions expressed at varying intensities using 12 scripted sentences. Each clip is crowd-rated in audio, visual, and audio-visual modalities. We use a random split with 6,698 samples for training and validation and 744 for testing.
\textbf{Kinetics-Sounds}\cite{Kinetics-Sounds1,Kinetics-Sounds2} is an audio-visual benchmark for action recognition in real-world videos, curated from Kinetics by selecting classes with distinctive sound. It contains 19,000 ten-second clips across 34 categories with official splits of 15,000 train, 1,900 validation, and 1,900 test, covering sound-producing actions such as musical performance, tool use, tap dancing, and singing. We use both audio and visual streams under the official split.
\textbf{CMU-MOSI}\cite{CMU-MOSI} is a multimodal dataset for fine-grained opinion-level sentiment analysis in online monologues, containing 2,199 clips from 93 videos by 89 speakers with aligned linguistic, visual, and acoustic streams. Most content consists of movie reviews, with transcripts, word-level alignments, and pre-extracted features provided. We use the official splits and all three modalities in our experiments.
\textbf{MUSIC-AVQA}\cite{MUSIC-AVQA} is a large-scale audio-visual dataset for question answering on music performance videos, with 9,288 videos and 45,867 question–answer pairs generated from 33 templates for spatio-temporal reasoning. It covers solos and ensembles across 22 instruments in four families and includes audio, visual, and audio-visual questions ranging from existence and location to counting, comparison, and temporal reasoning.
\textbf{UCF-ZSL}\cite{UCF,AVCA,STFT} is a zero-shot action recognition benchmark derived from the UCF101 real-world video corpus, containing 101 action categories and 13,320 videos, evaluated by separating classes into seen and unseen and linking videos to semantic label embeddings. We use the data processing method in \cite{AVCA} and report both zero-shot results in our experiments.

\subsection{Experimental Settings}
For data processing and the model backbone, we follow the feature processing strategy in InfoReg\cite{InfoReg}. For CREMA-D and Kinetics-Sounds, we use ResNet-18 as the backbone \cite{RESNET}. For CMU-MOSI, we use a Transformer-based model\cite{transformer}. All components are trained from scratch to ensure that feature extraction is well adapted to our specific tasks and datasets. For the MUSIC-AVQA dataset, we use three AVQA baseline models: AVSD\cite{AVSD}, PSTP\cite{PSTP}, and SHMamba\cite{SHMamba} to evaluate the performance of RedReg. For the AVCA task, we use three baselines: AVCA\cite{AVCA}, MSTR\cite{MSTR}, and STFT\cite{STFT} on the UCF-ZSL dataset. RedReg is trained on two 40 GB NVIDIA A100 GPUs. To remain consistent with prior work, we match its settings on CREMA-D, Kinetics-Sounds, CMU-MOSI, MUSIC-AVQA and UCF-ZSL. Specifically, we use SGD with momentum $0.9$ as the optimizer, an initial learning rate of $0.002$, and a batch size of $64$. The threshold $R$ is set to $0.15$. We set the hyperparameters as follows: $\beta=0.9$, $\gamma=0.5$, $\tau_{\min}=0.2$, and $\tau_{\max}=0.5$.

To ensure a fair comparison, we use the evaluation method proposed in \cite{InfoReg}. The evaluation metrics include accuracy and F1 score:
\begin{equation}
\mathrm{Acc}=\frac{\sum_{i=1}^B1_{\hat{y_i}=y_i}}{B},
\end{equation}

\begin{equation}
\mathrm{F1~score}=2\cdot\frac{Precision\cdot Recall}{Precision+Recall}.
\end{equation}
where $B$ is the number of testing samples, and $\hat{y_i}$ is the prediction of model for sample $x_i$ in testing set.

\begin{table*}
    \centering
    \begin{threeparttable}
        \caption{Comparison to imbalanced multimodal baselines, all optimized using one multimodal loss.The best and second best are bolded and underlined respectively.}
        \label{main1}
        \fontsize{10}{14}\selectfont % 调整字号大小为10，行距大小为12
        \setlength{\tabcolsep}{4pt} % 减少列之间的间隔，以适应更大的字体
        \begin{tabular}{c|ccc|ccc}
            \toprule[1.5pt]  
            \multirow{2}{*}{\textbf{Method}} &
            \multicolumn{3}{c|}{\begin{tabular}[c]{@{}c@{}}\textbf{CREMA-D}\end{tabular}} &
            \multicolumn{3}{c}{\begin{tabular}[c]{@{}c@{}}\textbf{Kinetics Sounds}\end{tabular}} \\ 
            & \textbf{Accuracy} & \textbf{Acc audio} & \textbf{Acc video}  & \textbf{Accuracy} & \textbf{Acc audio} & \textbf{Acc video}  \\ \hline        
         Joint trainning    & 66.47 & 58.31 & 38.92 & 64.27 & 53.83 & 36.62  \\ 
         OGM\cite{OGM}\textsuperscript{CVPR2022} & 69.11 & 57.35 & 39.24 & 67.38 & 54.15 & 40.82 \\
         Greedy\cite{Greedy}\textsuperscript{ICML2022}        & 68.81 & 58.37 & 40.13 & 67.12 & 54.02 & 37.95 \\
         PMR\cite{PMR}\textsuperscript{CVPR2024}  & 67.24 & 57.47 & 39.34 & 67.26 & 54.38 & 37.18 \\
         AGM\cite{AGM}\textsuperscript{ICCV2023}       & 70.14 & 59.87 & 45.16 & 67.11 & 54.67 & 38.94 \\
         DRBML\cite{DRBML}\textsuperscript{ECCV2024}        & 71.02 & \underline{61.84} & 47.84 & \underline{70.03} & 55.13 & 43.14 \\
         InfoReg\cite{InfoReg}\textsuperscript{CVPR2025}        & \underline{71.76} & 60.35 & \underline{48.27} & 69.77 & \underline{55.46} & \textbf{45.17} \\

         \hline
            % \rowcolor{gray!20}
         RedReg & \textbf{73.86} & \textbf{62.43} & \textbf{49.72} & \textbf{71.13} & \textbf{56.47} & \underline{44.69} \\

            \bottomrule[1.5pt]
        \end{tabular}
    \end{threeparttable}
\end{table*}

\begin{table*}
    \centering
    \begin{threeparttable}
        \caption{Compared with the unbalanced multimodal baseline, both unimodal and multimodal losses are used for optimization.The best and second best are bolded and underlined respectively}
        \label{main2}
        \fontsize{10}{14}\selectfont % 调整字号大小为10，行距大小为12
        \setlength{\tabcolsep}{4pt} % 减少列之间的间隔，以适应更大的字体
        \begin{tabular}{c|ccc|ccc}
            \toprule[1.5pt]  
            \multirow{2}{*}{\textbf{Method}} &
            \multicolumn{3}{c|}{\begin{tabular}[c]{@{}c@{}}\textbf{CREMA-D}\end{tabular}} &
            \multicolumn{3}{c}{\begin{tabular}[c]{@{}c@{}}\textbf{Kinetics Sounds}\end{tabular}} \\ 
            & \textbf{Accuracy} & \textbf{Acc audio} & \textbf{Acc video}  & \textbf{Accuracy} & \textbf{Acc audio} & \textbf{Acc video}  \\ \hline        
         Joint trainning    & 66.47 & 58.31 & 38.92 & 64.27 & 53.83 & 36.62  \\ 
         Joint trainning*   & 70.73 & 58.27 & 57.34 & 68.13 & 54.29 & 43.13 \\
         G-Blending\cite{G-Blending}\textsuperscript{CVPR2020}        & 69.62 & 60.83 & 54.38 & 68.39 & 55.27 & 43.82 \\
         MMPareto\cite{MMPareto}\textsuperscript{ICML2024}  & 73.24 & 60.31 & 59.17 & 71.12 & 55.73 & 53.26 \\
         InfoReg*\cite{InfoReg}\textsuperscript{CVPR2025}       & \underline{74.86} & \underline{62.14} & \underline{63.67}& \underline{71.86} & \underline{57.38} & \underline{54.11} \\
         \hline
            % \rowcolor{gray!20}
         RedReg* & \textbf{77.68} & \textbf{62.64} & \textbf{64.82} & \textbf{74.04} & \textbf{59.67} & \textbf{56.87} \\

            \bottomrule[1.5pt]
        \end{tabular}
    \end{threeparttable}
\end{table*}

\subsection{Results and Comparison}
We compare RedReg with OGM\cite{OGM}, Greedy\cite{Greedy}, PMR\cite{PMR}, AGM\cite{AGM}, InfoReg\cite{InfoReg}, G-Blending\cite{G-Blending}, MMPareto\cite{MMPareto}, and DRBML\cite{DRBML}. G-Blending, MMPareto, and InfoReg* optimize with both unimodal and multimodal losses, whereas OGM, Greedy, PMR, AGM, DRBML, and InfoReg rely only on unimodal losses. Joint training represents a general baseline for concatenated fusion using a single multimodal cross entropy. Joint training* uses a joint optimization involving unimodal and multimodal losses.

In the multimodal-only setting (Table \ref{main1}), RedReg achieves the highest overall accuracy on both datasets. It reaches 73.86\% on CREMA-D and 71.13\% on Kinetics-Sounds. These results outperform the best baseline methods by 2.10\% and 1.10\%. By modality, RedReg yields the top audio accuracy on both datasets and improves the video branch on CREMA-D while matching the strongest baseline on Kinetics-Sounds. Relative to Joint training, the audio–video gap shrinks markedly (from 19.39\% to 12.71\% on CREMA-D and from 17.21\% to 11.78\% on Kinetics-Sounds), indicating a more balanced contribution from the two streams. These gains are consistent with our design: redundancy regulation suppresses repeatedly acquired signals from the dominant modality and reallocates gradient budget to complementary cues, improving performance even without unimodal supervision. When unimodal losses are added (Table \ref{main2}), all methods improve, confirming that unimodal supervision helps each stream extract useful evidence. RedReg* remains best on both datasets with 77.68\% on CREMA-D and 74.04\% on Kinetics-Sounds, and delivers the strongest video improvements while further reducing the modality gap (to 2.18\% and 2.80\%). The results demonstrate that our redundancy-aware regularizer works synergistically with the unimodal objective.
The model prioritizes non-overlapping information, preventing the dominant modality from absorbing too much training signal while preserving the learning power of the weaker modality.
\begin{table*}
    \centering
    \begin{threeparttable}
        \caption{Ablation for RedReg on CREMA-D datasets, all optimized using one multimodal loss. The best is bolded.}
        \label{ablation1}
        \fontsize{10}{14}\selectfont % 第一个数字是字号大小，第二个数字是行距大小
        \setlength{\tabcolsep}{5pt}{
            \begin{tabular}{c|ccc|ccc}
                \toprule[1.5pt]  
                \multirow{2}{*}{\textbf{Model}} &
                  \multicolumn{3}{c|}{\begin{tabular}[c]{@{}c@{}}\textbf{CREMA-D}\end{tabular}} &
                  \multicolumn{3}{c}{\begin{tabular}[c]{@{}c@{}}\textbf{Kinetics Sounds}\end{tabular}} \\
                & \textbf{Accuracy}     & \textbf{Acc audio}    & \textbf{Acc video}     & \textbf{Accuracy}     & \textbf{Acc audio}     & \textbf{Acc video}         \\ \hline
                Joint training     & 66.47 & 58.31 & 38.92 & 64.27 & 53.83 & 36.62  \\
                W/o Redundant Phase Monitor      & 69.51 & 59.37 & 44.28  & 68.43 & 54.16 & 39.17   \\ 
                W/o Co-info Gating        & 71.37  & 60.72 & 47.31 & 69.38 & 54.81 & 42.17          \\\hline
                     % \rowcolor{gray!20}
                RedReg        & \textbf{73.86} & \textbf{62.43} & \textbf{49.72} & \textbf{71.13} & \textbf{56.47} & \textbf{44.69}  \\ 
                  \bottomrule[1.5pt]
            \end{tabular}
        }
    \end{threeparttable}
\end{table*}

\begin{table*}
    \centering
    \begin{threeparttable}
        \caption{Ablation for RedReg on CREMA-D datasets, both unimodal and multimodal losses are used for optimization. The best is bolded.}
        \label{ablation2}
        \fontsize{10}{14}\selectfont % 第一个数字是字号大小，第二个数字是行距大小
        \setlength{\tabcolsep}{5pt}{
            \begin{tabular}{c|ccc|ccc}
                \toprule[1.5pt]  
                \multirow{2}{*}{\textbf{Model}} &
                  \multicolumn{3}{c|}{\begin{tabular}[c]{@{}c@{}}\textbf{CREMA-D}\end{tabular}} &
                  \multicolumn{3}{c}{\begin{tabular}[c]{@{}c@{}}\textbf{Kinetics Sounds}\end{tabular}} \\
                & \textbf{Accuracy}     & \textbf{Acc audio}    & \textbf{Acc video}     & \textbf{Accuracy}     & \textbf{Acc audio}     & \textbf{Acc video}         \\ \hline
                Joint training*     & 70.73 & 58.27 & 57.34 & 68.13 & 54.29 & 43.13  \\
                W/o Redundant Phase Monitor      & 73.42 & 59.41 & 60.15  & 70.23 & 55.11 & 48.07   \\ 
                W/o Co-info Gating        & 75.09  & 61.33 & 61.81 & 73.16 & 55.34 & 54.38          \\\hline
                     % \rowcolor{gray!20}
                RedReg*        & \textbf{77.68} & \textbf{62.64} & \textbf{64.82} & \textbf{74.04} & \textbf{59.67} & \textbf{56.87}  \\ 
                  \bottomrule[1.5pt]
            \end{tabular}
        }
    \end{threeparttable}
\end{table*}
\subsection{Redundancy Analysis}
As shown in the Fig. \ref{ER}, we evaluate RedReg on CREMA-D using three training-time metrics. RLC measures the strength of representation-to-logit coupling. Higher RLC means a more decisive, discriminative contribution. DGR (effective gain growth rate) tracks how rapidly a modality acquires new, non-overlapping information. RS is the redundancy score and reflects duplicated or overlapping signals. Lower RS indicates less redundancy and more efficient gradient use. 

Compared with the baseline, RedReg shifts all three indicators in the desired direction. The baseline shows diverging RLC: audio dominates early and stabilizes, while visual rises late with high variance. DGR decays rapidly toward a near-zero plateau, and RS briefly drops before rebounding and settling high, revealing persistent redundancy and suppressed visual gain. With RedReg, audio and visual RLC grow smoothly to similar levels, indicating balanced and decisive contributions. Visual-stream DGR rises steadily and matches the audio stream near the end of the prime stage, indicating faster acquisition of new information. RS decreases monotonically to a lower plateau and remains low, confirming sustained suppression of redundant updates. All curves also exhibit lower variance, indicating a more stable optimization trajectory.This is mainly attributed to the fact that RedReg prevents the over-coupling of the dominant modes and reallocates the gradient budget to the weaker modes.

\begin{table}
    \centering
    \begin{threeparttable}
        \caption{Comparison of different fusion Method on CREMA-D datasets. The best is bolded.}
        \label{fusion}
        \fontsize{10}{12}\selectfont % 调整字号大小为10，行距大小为12
        \setlength{\tabcolsep}{3pt} % 减少列之间的间隔，以适应更大的字体
        \begin{tabular}{c|ccc}
            \toprule[1.5pt]  
            \multirow{2}{*}{\textbf{Method}} &
            \multicolumn{3}{c}{\begin{tabular}[c]{@{}c@{}}\textbf{CREMA-D}\end{tabular}} \\ 
            & \textbf{Accuracy} & \textbf{Acc audio} & \textbf{Acc video} \\ \hline        
         Concatenation          & 66.47 & 58.31 & 38.92   \\ 
         Concatenation-InfoReg  & 71.76 & 60.35 & 48.27  \\
                     % \rowcolor{gray!20}
         Concatenation-RedReg   & \textbf{73.86} & \textbf{62.43} & \textbf{49.72}   \\
         \hline         
         Summation         & 64.38 & 56.27 & 34.18  \\
         Summation-InfoReg & 70.12 & 59.36 & 46.21  \\
                     % \rowcolor{gray!20}
         Summation-RedReg  & \textbf{71.38} & \textbf{60.23} & \textbf{47.58}  \\
           \hline
         FiLM              & 66.67 & 57.28 & 39.52  \\
         FiLM-InfoReg      & 70.23 & \textbf{59.49} & 48.52  \\
                     % \rowcolor{gray!20}
         FiLM-RedReg       & \textbf{71.74} & 59.34 & \textbf{49.73} \\
         \hline
         Gated             & 65.32 & 54.17 & 36.48  \\
         Gated-InfoReg     & 69.18 & 58.24 & 47.36  \\
                     % \rowcolor{gray!20}
         Gated-RedReg      & \textbf{70.25} & \textbf{57.38} & \textbf{48.87}  \\

            \bottomrule[1.5pt]
        \end{tabular}
    \end{threeparttable}
\end{table}
\subsection{Ablation Study}
This section uses ablation studies to explore the role of each key component in the RedReg method and its effectiveness under different fusion methods.
\subsubsection{Effectiveness of RedReg Components}
We analyze two key components of RedReg: the redundant phase monitor and the co-information gating mechanism. We present results under two settings: using only multimodal loss, and using both unimodal and multimodal losses (Tables \ref{ablation1} and \ref{ablation2}). In the multimodal-only setting, removing either component degrades performance and increases the gap between audio and video. The performance drop is more significant for the video branch. This result reveals two main effects.  First, without the redundant phase monitor, the dominant stream repeats similar signals, reducing the useful gradients for the weaker stream. Second, without the co-information gate, the model transmits many redundant updates, leading to underuse of complementary cues. When both components are included, the model performs better, and the two streams contribute more equally. When both components are included, the model performs better, and the two streams contribute more equally.

In the joint-loss setting, all variants show improvement, but the same pattern remains. The model without the redundant phase monitor still exhibits audio dominance and limited benefit for video. The model without the co-information gate also performs worse due to unfiltered overlapping information. The complete RedReg* model achieves the best and most balanced performance.  This was attributed to several metrics: a more balanced representation coupled with logistic regression, more persistent gains in the visual stream, and lower redundancy. In summary, the redundant phase monitor determines when to suppress repeated signals, while the co-information gating mechanism decides what information to transmit. The combined effect of the two alleviates the modal imbalance.

\begin{figure*}
  \centering
  \subfloat[$\gamma$]{\includegraphics[width=0.32\linewidth]{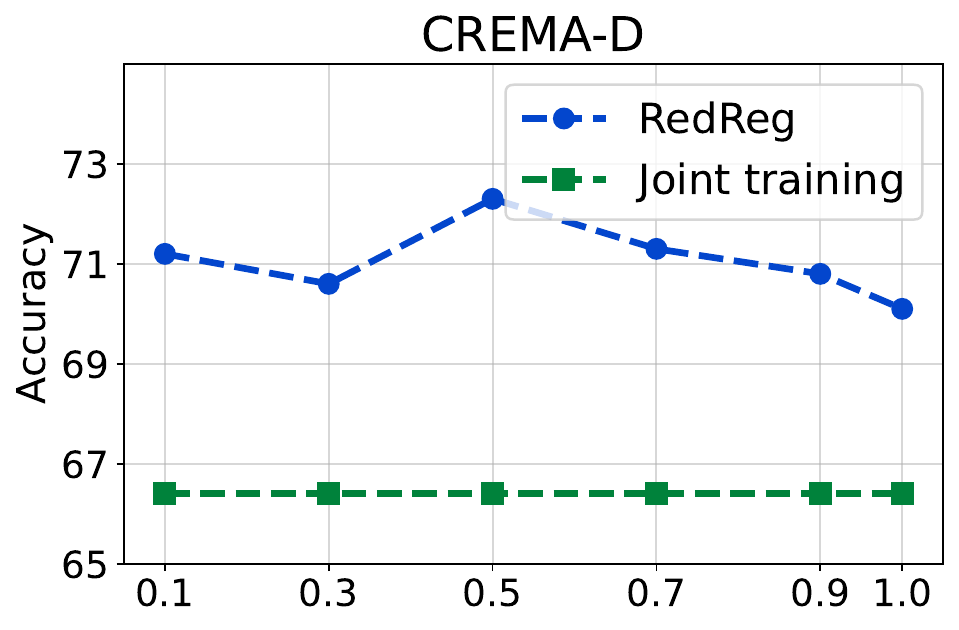}\label{H(a)}}
  \hfill % 控制子图之间的空间
  \subfloat[$\beta$]{\includegraphics[width=0.32\linewidth]{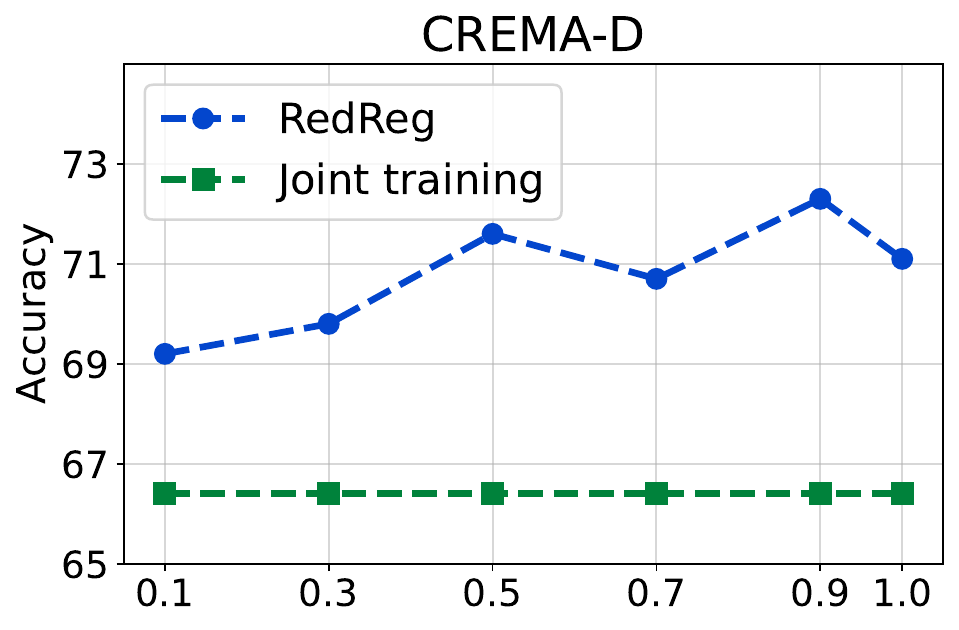}\label{H(b)}}
  \subfloat[$R$]{\includegraphics[width=0.32\linewidth]{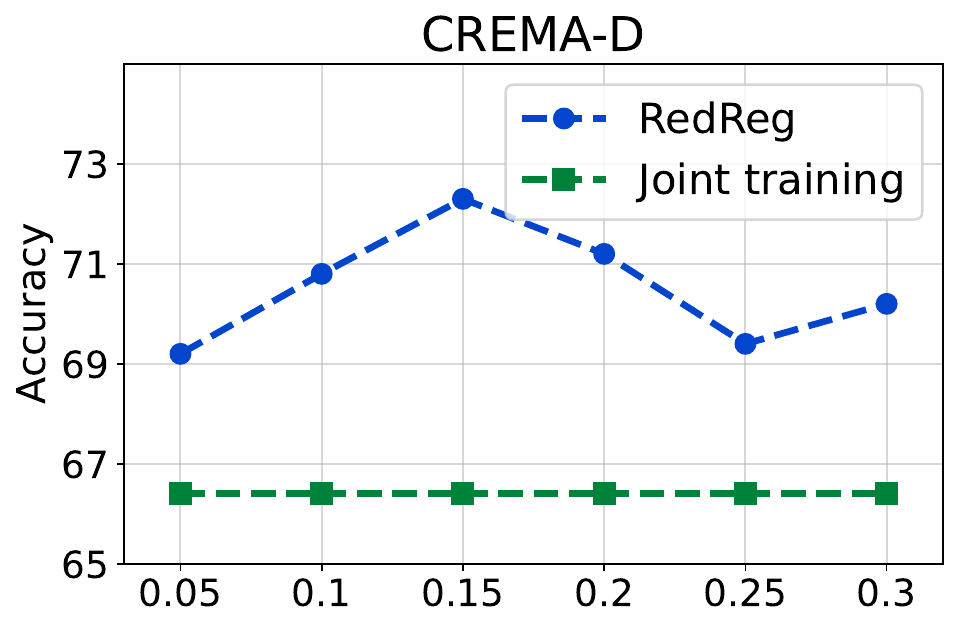}\label{H(c)}}
  \caption{Hyperparameter experimental analysis on CREMA-D using a single multimodal cross-entropy loss: \textbf{(a).} The overall accuracy of different $\gamma$. \textbf{(b).} The overall accuracy of different $\beta$. \textbf{(c).} The overall accuracy of different $R$.}
  \label{O-HY}
\end{figure*}

\begin{figure*}
  \centering
  \subfloat[$\gamma$*]{\includegraphics[width=0.32\linewidth]{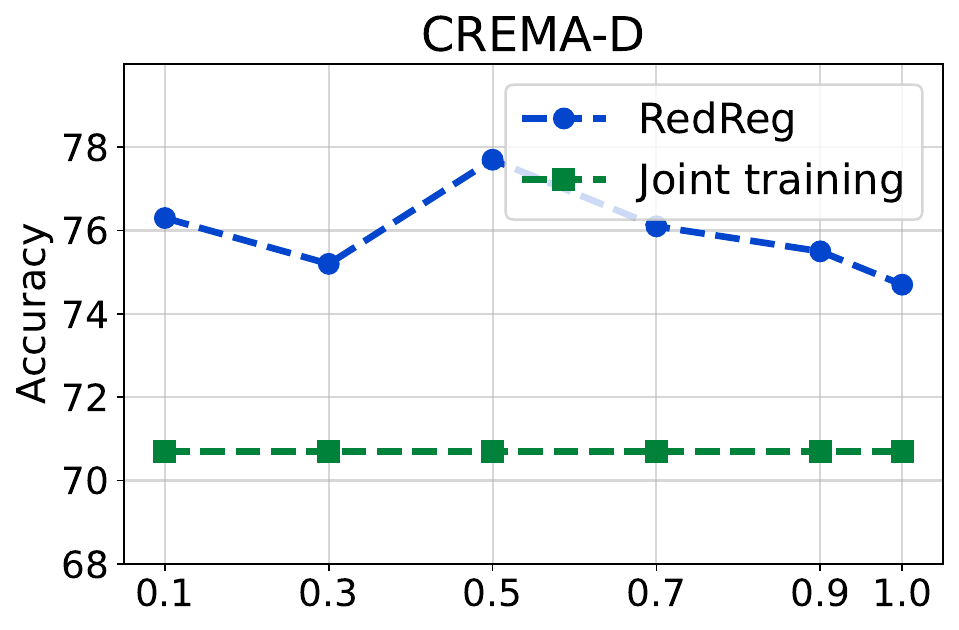}\label{HY(a)}}
  \hfill % 控制子图之间的空间
  \subfloat[$\beta$*]{\includegraphics[width=0.32\linewidth]{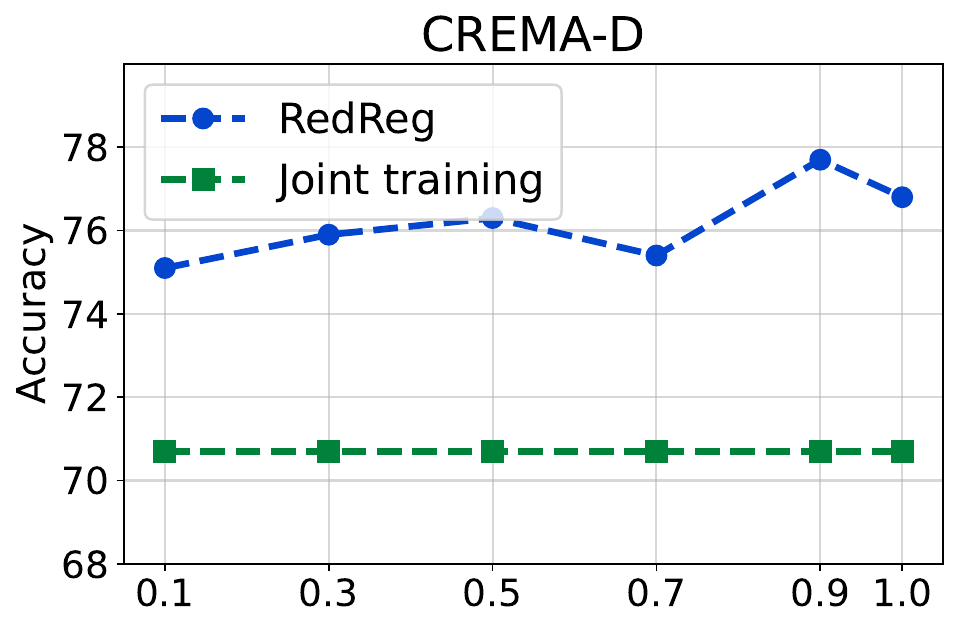}\label{HY(b)}}
  \subfloat[$R$*]{\includegraphics[width=0.32\linewidth]{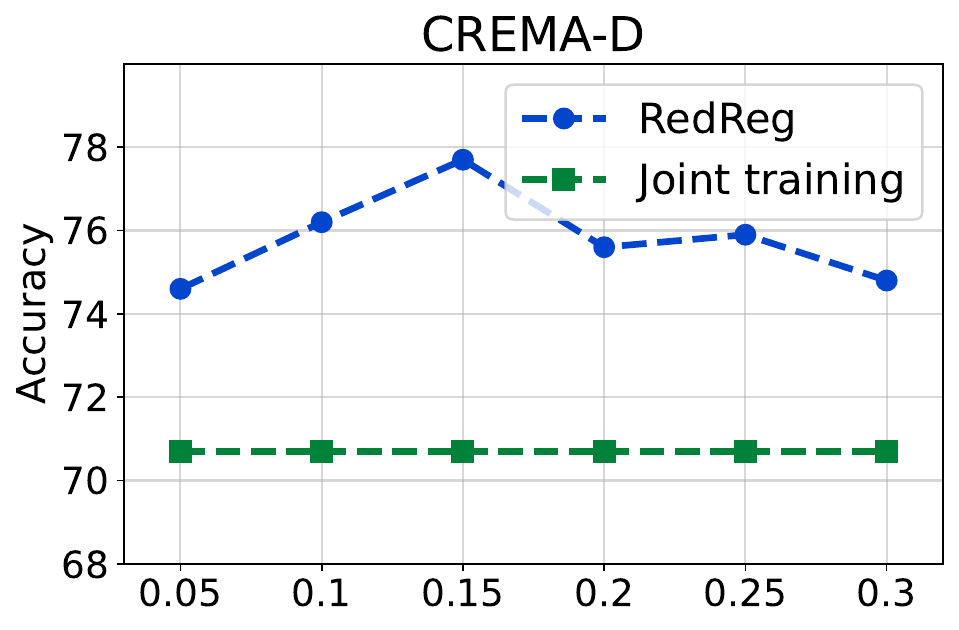}\label{HY(c)}}
  \caption{Hyperparameter experimental analysis on CREMA-D using both unimodal and multimodal losses: \textbf{(a).} The overall accuracy of different $\gamma$*. \textbf{(b).} The overall accuracy of different $\beta$*. \textbf{(c).} The overall accuracy of different $R$*.}
  \label{Z-HY}
\end{figure*}

\subsubsection{Combination with Different Fusion Methods}
We evaluate four fusion methods on CREMA-D in Table \ref{fusion}, namely Concatenation, Summation, FiLM\cite{FILM}, and Gated fusion\cite{GATE}. RedReg operates only during training and leaves the inference-time fusion unchanged. With concatenation, overall accuracy increases from 66.47\% to 73.86\%, with 62.43\% audio and 49.72\% video. With summation, overall accuracy increases from 64.38\% to 71.38\%, with 60.23\% audio and 47.58\% video. With FiLM, overall accuracy increases from 66.67\% to 71.74\%, with 59.34\% audio and 49.73\% video. With gated fusion, overall accuracy increases from 65.32\% to 70.25\%, with 57.38\% audio and 48.87\% video. 

RedReg outperforms InfoReg under all fusion methods, likely due to the preservation of modality-specific information by co-information gating mechanism. Overall accuracy improves by 2.10\% with concatenation, 1.26\% with summation, 1.51\% with FiLM, and 1.07\% with gated fusion. Video accuracy increases by 1.21\% to 1.46\% over InfoReg, while audio accuracy stays similar. These results align with the design. The redundant phase monitor reduces repeated signals within the prime learning window and frees gradient budget for the weaker stream. The co-information gate preserves complementary signals throughout training. The method is compatible with additive fusion, feature-wise affine fusion such as FiLM, and gated fusion. The effect remains stable across fusion choices, and both accuracy and balance improve.

\begin{figure*}
  \centering
  \subfloat[Joint training]{\includegraphics[width=0.24\linewidth,clip,trim=2pt 2pt 2pt 2pt]{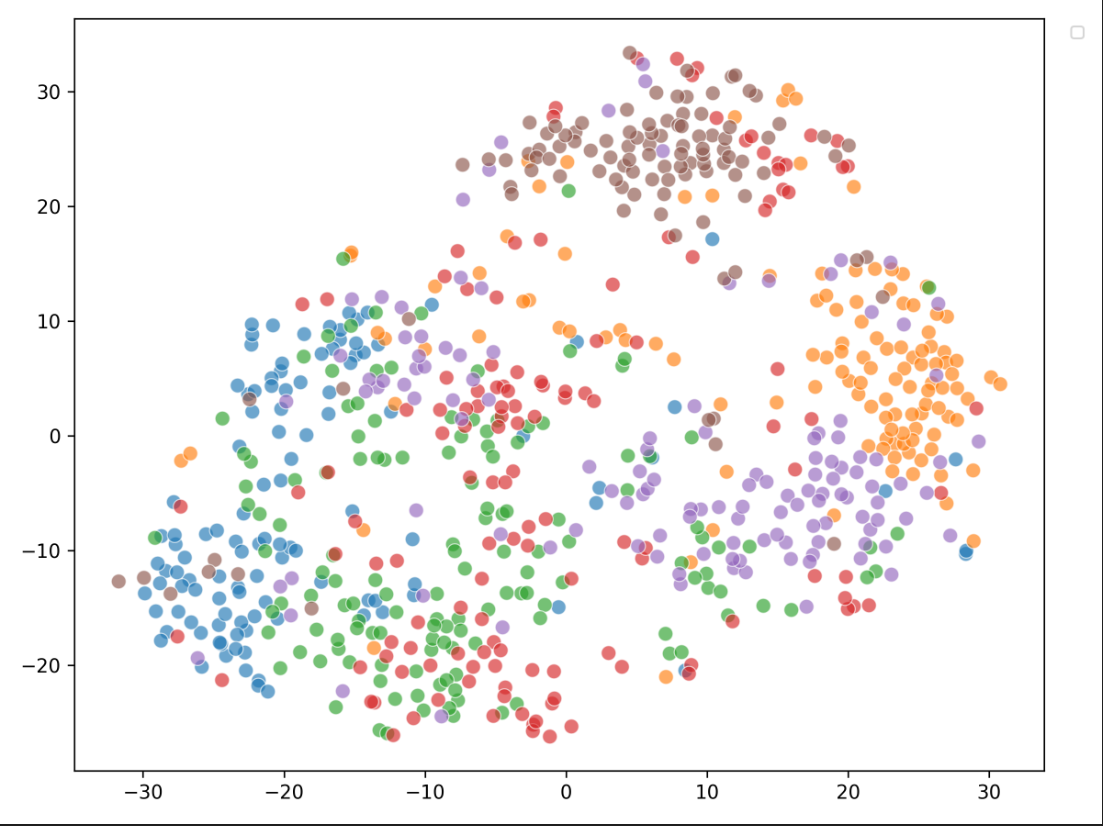}\label{tsneO}}
  \subfloat[RedReg]{\includegraphics[width=0.24\linewidth]{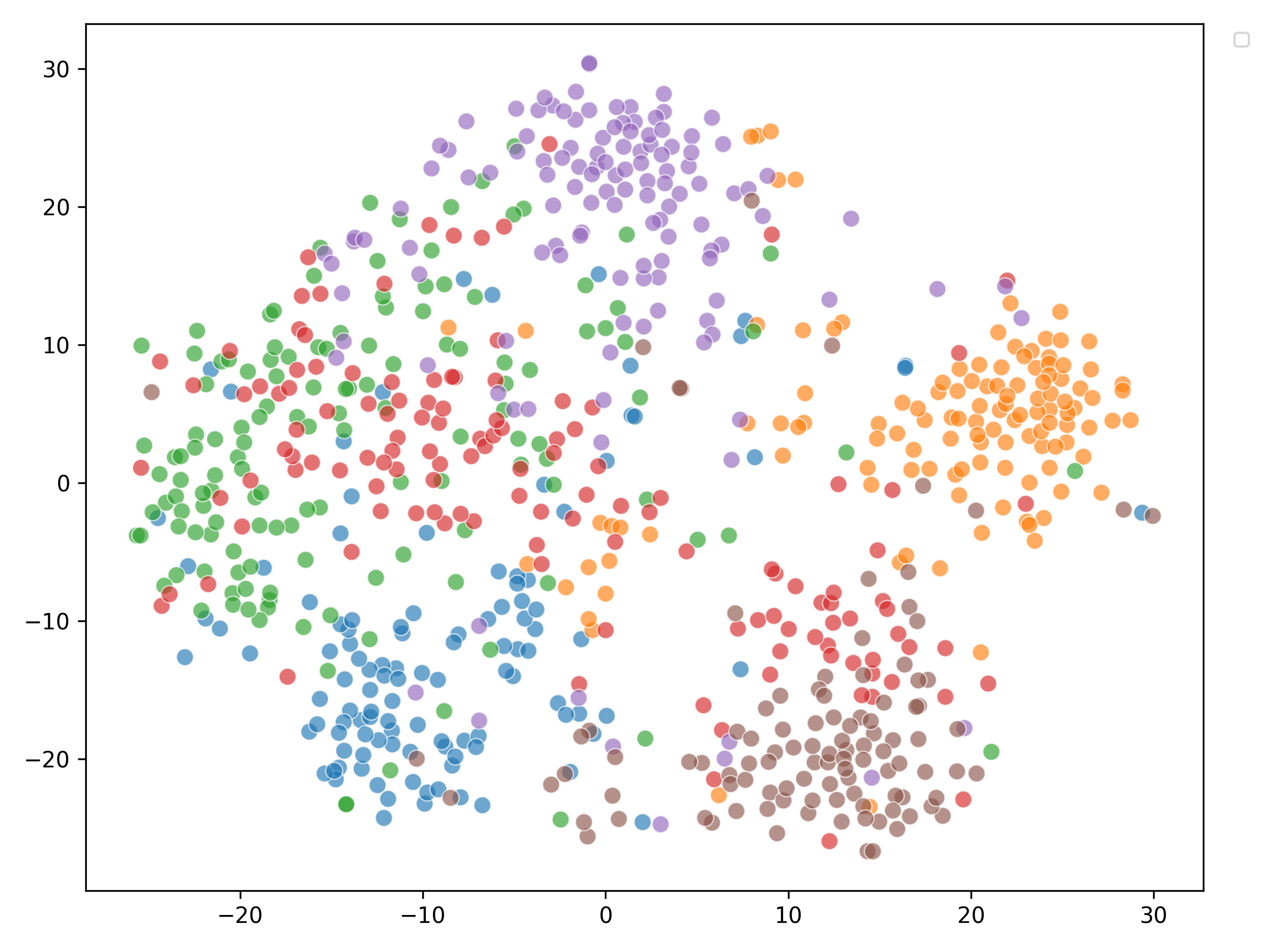}\label{tsneZ}}
  \subfloat[Joint training*]{\includegraphics[width=0.24\linewidth]{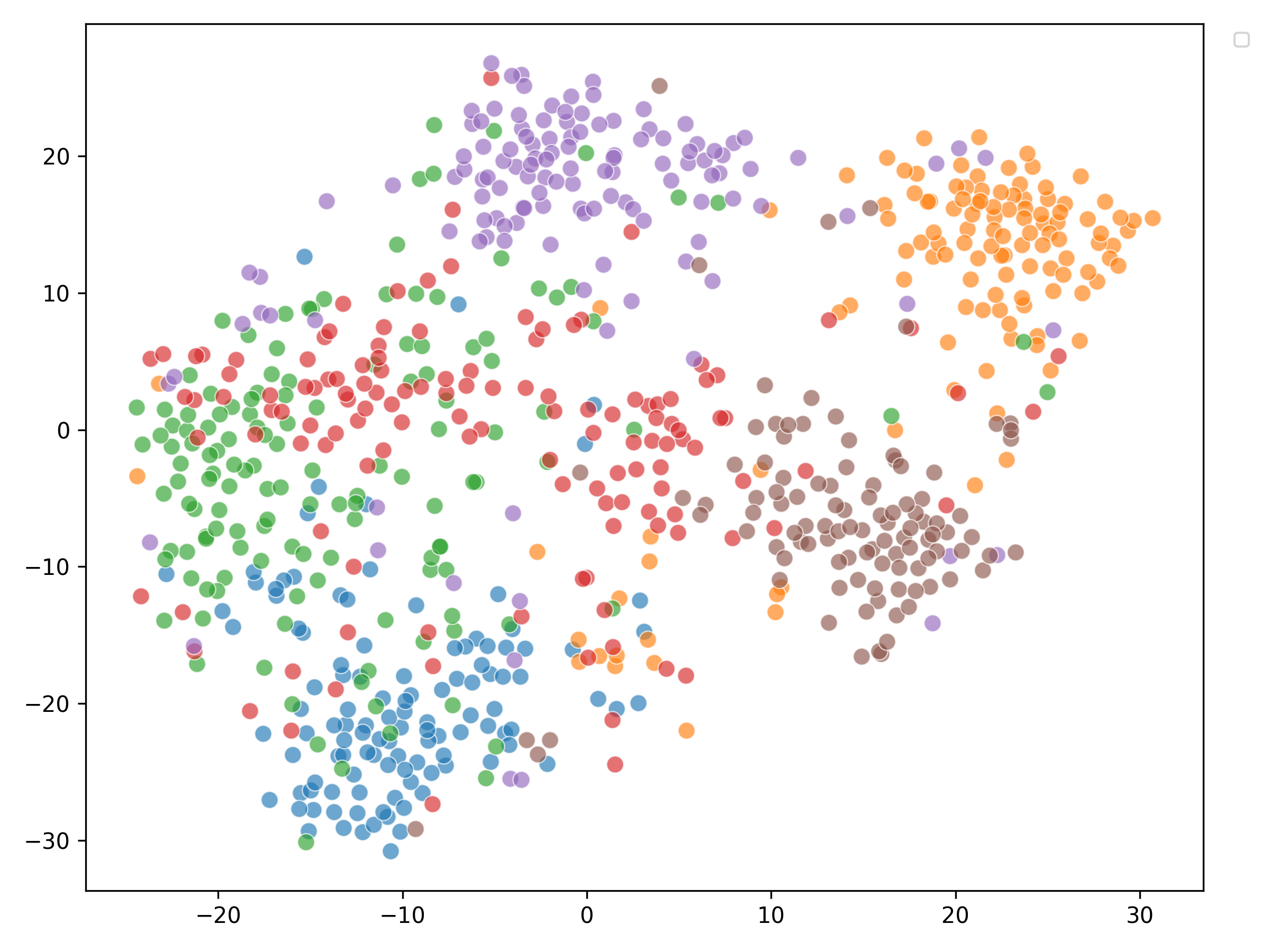}\label{tsneooo}}
  \subfloat[RedReg*]{\includegraphics[width=0.24\linewidth]{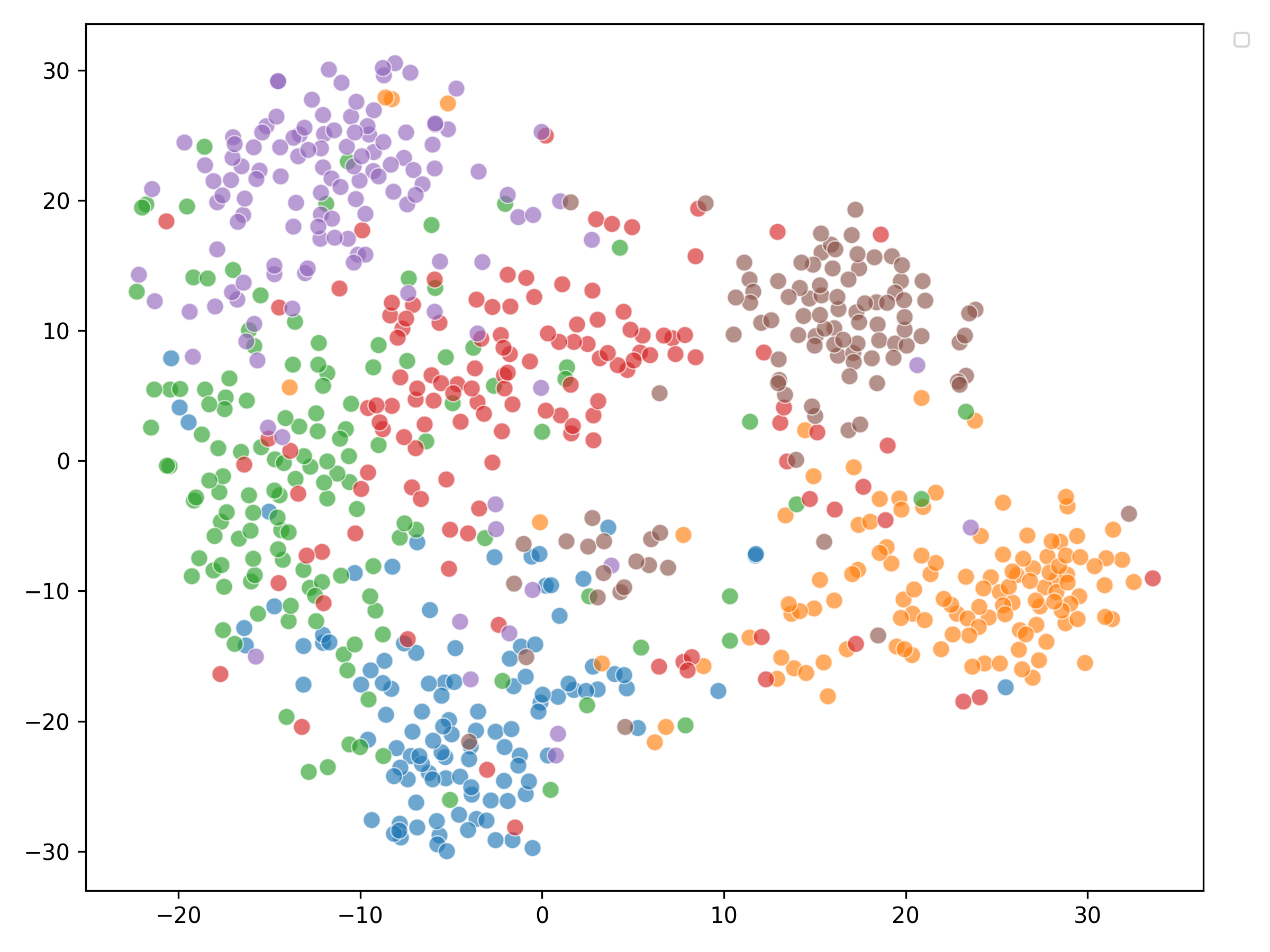}\label{tsnezzz}}
  \caption{The figure shows the representation of multimodal features on CREMA-D in RedReg and the baseline under different training losses using t-SNE.}
  \label{tsne}
\end{figure*}

\subsection{Hyper Parameter Setting}
\subsubsection{Influence Coefficient $\gamma$}From Fig. \ref{H(a)} and Fig. \ref{HY(a)}, when using a single multimodal cross-entropy loss, accuracy first increases and then decreases as $\gamma$ grows, peaking at a moderate value around 0.5. Further increases make the penalty overly strong, suppressing useful redundant signals and causing performance to drop. Under the joint unimodal-plus-multimodal loss setting, the overall accuracy rises further and the curve becomes smoother. This indicates that a moderate sparsity penalty maximizes the discriminative power of the redundancy indicator, since too small fails to suppress noise, while too large suppresses beneficial cues.

\subsubsection{Smoothing Coefficient $\beta$}Fig. \ref{H(b)} and Fig. \ref{HY(b)} show the same middle-is-best pattern. When using only the multimodal loss, a too small $\beta$ makes the time series estimate overly sensitive to transient perturbations, introducing volatility. A too large $\beta$ can lead to significant lags, obscuring short-term discriminative information. A moderate $\beta$ (around 0.5) yields the best results. Using the joint loss, the curves show less volatility and a wider optimal range. Multi-objective constraints thus improve robustness to the smoothing strength, though moderate smoothing remains most effective.

\subsubsection{Gating Threshold $R$}Fig. \ref{H(c)} and Fig. \ref{HY(c)} show that $R$ sets the minimum redundancy indicator required to activate the redundant phase monitor. When using only the multimodal loss, a too low threshold leads to noise-driven triggering, while a too high threshold filters out valid redundancy. Both extremes impair performance. Under the joint loss, overall accuracy improves further, with the optimal value remaining in the mid-range $R$* region, while performance degrades more significantly at the extremes. These results demonstrate that an appropriate minimum redundancy requirement can avoid spurious activations while preserving critical discriminative information.
\begin{table}
    \centering
    \begin{threeparttable}
        \caption{The performance comparison on CMU-MOSI datasets. The best and second best are bolded and underlined respectively.}
        \label{cmu}
        \fontsize{10}{12}\selectfont % 调整字号大小为10，行距大小为12
        \setlength{\tabcolsep}{15pt} % 减少列之间的间隔，以适应更大的字体
        \begin{tabular}{c|cc}
            \toprule[1.5pt]  
            \multirow{2}{*}{\textbf{Method}} &
            \multicolumn{2}{c}{\begin{tabular}[c]{@{}c@{}}\textbf{CMU-MOSI}\end{tabular}} \\ 
            & \textbf{Accuracy} & \textbf{F1 score}  \\    
                     \hline
         Joint training & 61.09 & 60.74  \\
         OGM         & 61.88 & 61.32  \\
         PMR         & 61.47 & 60.98 \\
         AGM         & 61.39 & 60.43   \\
         InfoReg     & \underline{62.31} & \underline{62.03}  \\
         \hline
                     % \rowcolor{gray!20}
         RedReg      & \textbf{62.76} & \textbf{63.41}   \\

            \bottomrule[1.5pt]
        \end{tabular}
    \end{threeparttable}
\end{table}

\subsection{Effectiveness in Complex Multimodal Task Scenarios}
\subsubsection{Complex Multimodal Model Backbone}
We evaluated RedReg on the trimodal CMU-MOSI dataset using a Transformer backbone model from \cite{transformer}. All methods used the same backbone model and training settings. As shown in Table \ref{cmu} , RedReg achieved the best results: 62.76\% accuracy and 63.41\% F1 score. Its accuracy surpassed that of joint training, OGM, PMR, AGM, and InfoReg. This improvement was observed in more complex settings. These findings demonstrate that RedReg scales well to complex model backbones.

\subsubsection{Complex Multimodal Tasks}
As shown in Table \ref{task}, we evaluate RedReg on two more challenging tasks. These are audio-visual zero-shot learning using the UCF-ZSL dataset and audio-visual question answering using the MUSIC-AVQA dataset. We apply RedReg to three backbone models for each task. The model structure and inference process remain unchanged. On UCF-ZSL, RedReg improves the performance of AVCA and STFT. The best result is achieved by STFT combined with RedReg. The performance of MSTR slightly decreases. This may be because the model already has strong alignment, so there is less space for improvement. 

On MUSIC-AVQA, RedReg improves the performance of AVSD and SHMamba. The highest score is obtained with SHMamba combined with RedReg. The performance of PSTP slightly decreases. This may result from the complexity of audio-visual question answering, where spatio-temporal interactions overlap with the function of our redundancy module. Overall, RedReg improves results in four out of six cases. The results show a clear trend. Redundancy control balances the two modalities and improves performance in transfer and reasoning tasks beyond basic classification.

\begin{table}
  \centering
  \begin{threeparttable}
    \caption{The performance comparison on UCF-ZSL and MUSIC-AVQA. The best is bolded}
    \label{task}
    \fontsize{10}{12}\selectfont
    \setlength{\tabcolsep}{5pt}
    \begin{tabular}{ccc|ccc}
      \toprule[1.5pt]
      \multicolumn{3}{c|}{\textbf{AVCA}} &
      \multicolumn{3}{c}{\textbf{AVQA}} \\
      \midrule
      \textbf{Method} &
      \multicolumn{1}{|c}{\textbf{w/o}} & \textbf{RedReg} &
      \textbf{Method} &
      \multicolumn{1}{|c}{\textbf{w/o}} & \textbf{RedReg} \\
      \midrule
      AVCA   & \multicolumn{1}{|c}{27.15} & \textbf{29.41} &
      AVSD    & \multicolumn{1}{|c}{66.90} & \textbf{67.83} \\
      MSTR   & \multicolumn{1}{|c}{\textbf{32.43}} & 32.18 &
      PSTP    & \multicolumn{1}{|c}{\textbf{73.42}} & 72.72 \\
      STFT   & \multicolumn{1}{|c}{32.58} & \textbf{33.25} &
      SHMamba & \multicolumn{1}{|c}{75.87} & \textbf{76.45} \\
      \bottomrule[1.5pt]
    \end{tabular}
  \end{threeparttable}
\end{table}
\subsection{Visualization}
To further assess the value of reducing information redundancy, we visualize audio–video features on CREMA-D using t-SNE\cite{TSNE}. We concatenate the two modalities and compare training strategies, as shown in Fig. \ref{tsne}. With multimodal loss only, the joint training baseline exhibits substantial class overlap and diffuse clusters. Many points lie near cluster boundaries, suggesting reliance on the dominant modality and continued learning of similar cues. With RedReg, clusters become tighter and inter-class gaps increase, and overlap is substantially reduced. This indicates that redundancy control enables the two modalities to provide complementary information and yields more discriminative features.

Under joint unimodal and multimodal losses, both methods improve, and RedReg* performs best. It shows larger inter-class separation and higher intra-class consistency. Clusters are clearer, outliers are fewer, and decision boundaries appear cleaner. These plots are consistent with the accuracy gains reported in Tables 1 and 2. This further supports the idea that reducing redundancy and guiding complementary cross-modal learning are key to achieving stronger multimodal representations and better overall performance.

\section{Limitations}
\label{E}
 
Although our method has achieved advantages in solving information redundancy, it still faces some limitations. We mainly focus on late-stage redundancy control and do not include the early critical learning window used in InfoReg. The two methods are potentially complementary. Early regulation can prevent the advantaged modality from dominating too quickly. Late-stage control can reduce ineffective updates and limit representation drift. A unified method would require robust estimation of modality sharing, time-aware thresholds and schedules, and careful handling of curriculum settings and hyperparameter coupling. In future work, we will explore integrated adaptive scheduling of early and late stages to improve the ability to discriminate collaborative information across different time periods.

\section{Conclusion}
\label{D}
This paper introduced RedReg, an online redundancy regulation strategy for multimodal learning that targets late-stage imbalance and redundancy. The method detects a redundancy stage with observable learning signals, gates on shared cross-modal semantics, and applies rate-limited updates in the subspace orthogonal to the task gradient. It needs no extra supervision or specialized modules and can be integrated into common backbones and fusion schemes. Across benchmarks, RedReg improves accuracy, validating the effectiveness.

\bibliographystyle{unsrt}
\bibliography{refs}

\vfill
\begin{IEEEbiography}[{\includegraphics[width=1in,height=1.25in,clip,keepaspectratio]{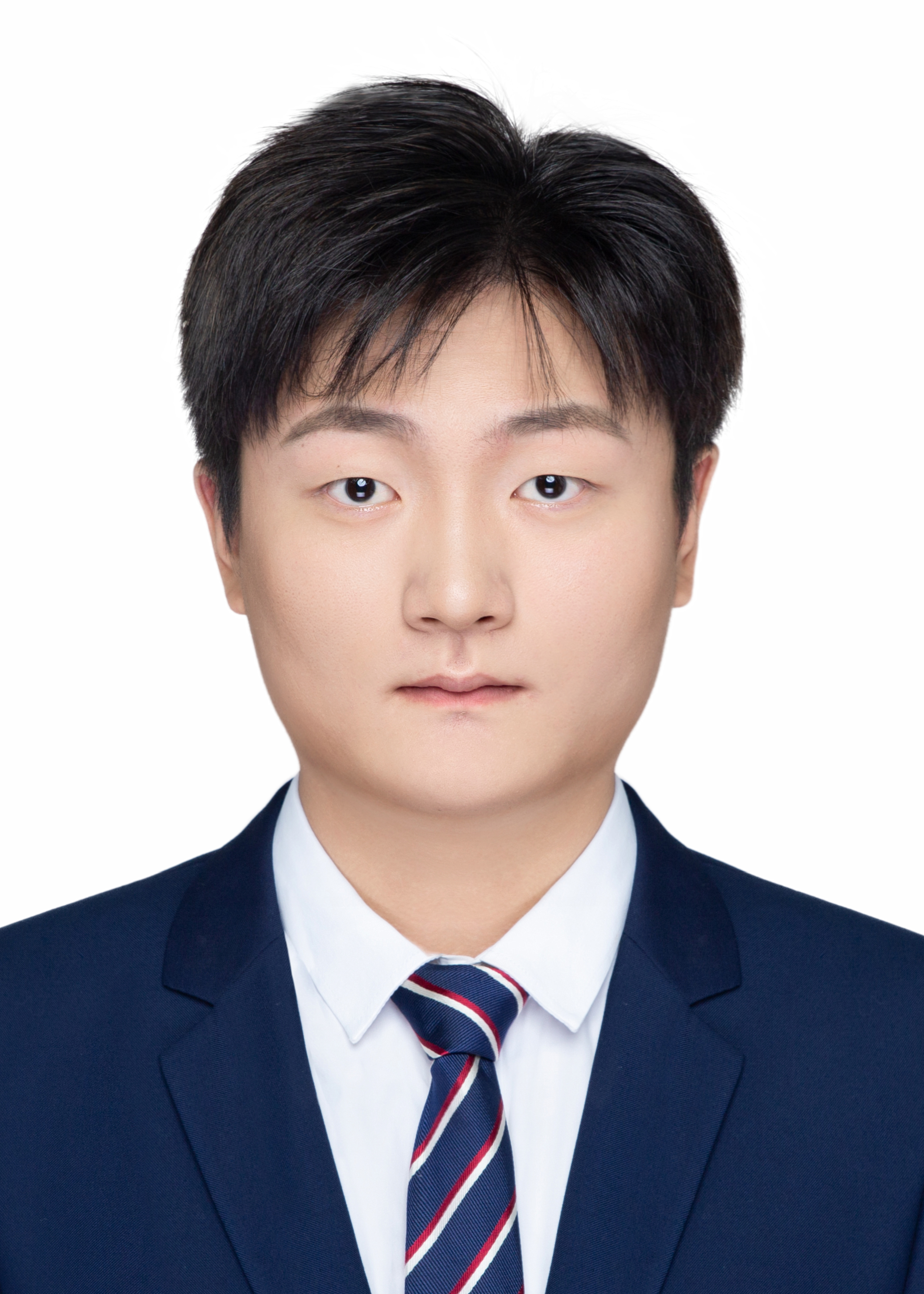}}]{Zhe Yang}  received the M.E. degree from the University of Electronic Science and Technology of China (UESTC) and is currently pursuing the Ph.D. degree with Harbin Institute of Technology (HIT). His research interests include zero-shot learning, audio-visual learning, and computer vision.
\end{IEEEbiography}

\begin{IEEEbiography}[{\includegraphics[width=1in,height=1.25in,clip,keepaspectratio]{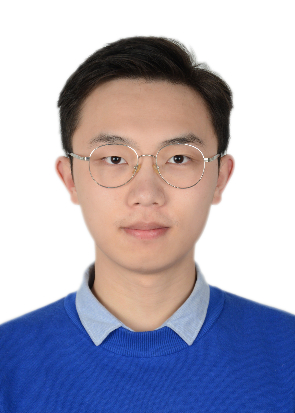}}]{Wenrui Li} received the B.S. degree from the School of Information and Software Engineering, University of Electronic Science and Technology of China (UESTC), Chengdu, China, in 2021. He is currently pursuing a Ph.D. degree in the School of Computer Science at Harbin Institute of Technology (HIT), Harbin, China. His research interests include multimedia search, joint source-channel coding, and spiking neural networks. He was supported by the National Natural Science Foundation of China (NSFC) Youth Student Basic Research Program (Doctoral Student) in 2024. He has authored or co-authored more than 30 technical articles in refereed international journals and conferences. 
\end{IEEEbiography}

\begin{IEEEbiography}[{\includegraphics[width=1in,height=1.25in,clip,keepaspectratio]{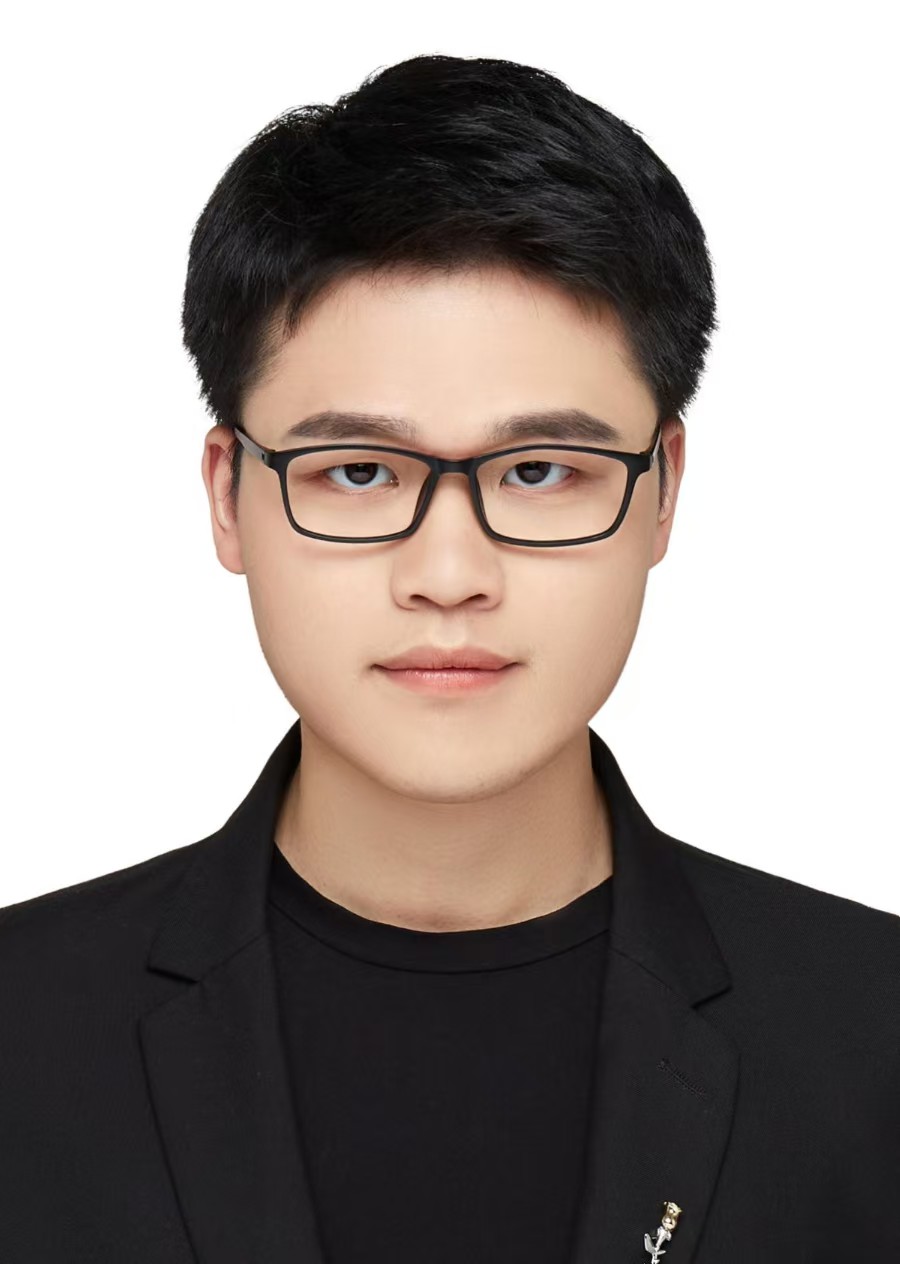}}]{Hongtao Chen} received the B.S. degree from the School of Science, Hangzhou Dianzi University (HDU), Hangzhou, China, in 2023. He is currently pursuing the M.S. degree with the School of Mathematical Sciences, University of Electronic Science and Technology of China (UESTC), Chengdu, China. His research interests include tensor completion, scientific computing, and computer vision.
\end{IEEEbiography}

\begin{IEEEbiography}[{\includegraphics[width=1in,height=1.25in,clip,keepaspectratio]{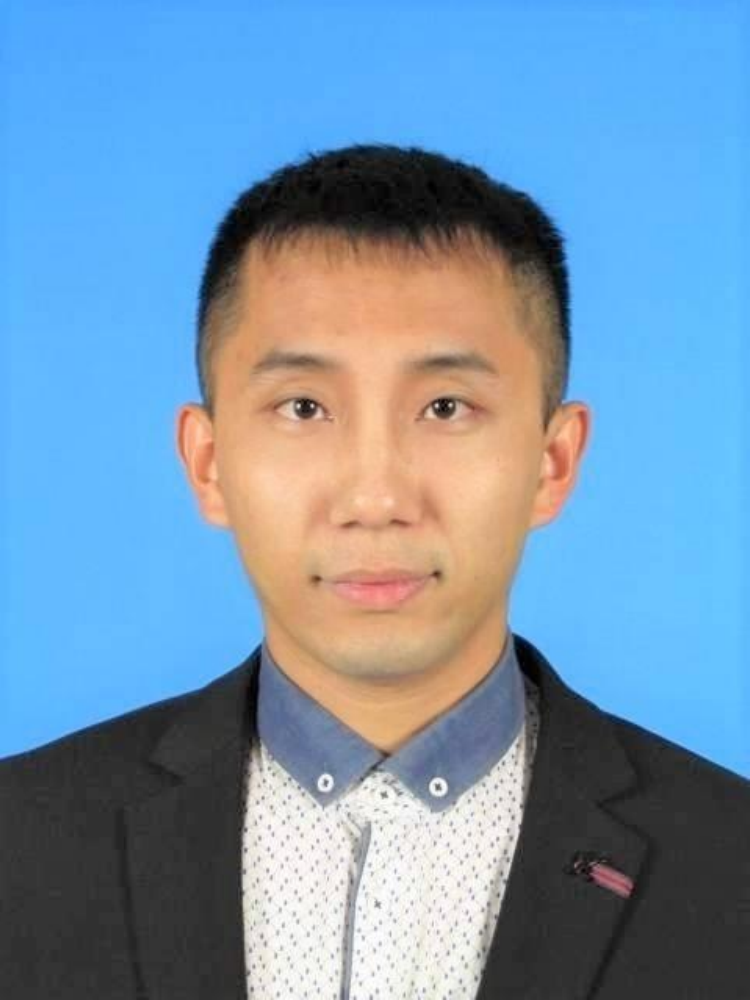}}]{Penghong Wang} received the M.S. degrees from Computer Science and Technology,  Taiyuan University of Science and Technology, Taiyuan, China, in 2020, and the Ph.D. degree in computer science from Harbin Institute of Technology (HIT), Harbin, China, in 2024. 
From 2021 to 2023, he was with Peng Cheng Laboratory, Shenzhen, China. He is currently an Associate Researcher with the Suzhou Institute of Advanced Research, HIT. 
His main research interests include  wireless sensor network, distributed source-channel coding and computer vision. 
He has authored or coauthored more than 20 technical papers in refereed international journals and conferences. 
His main research interests include wireless sensor networks, joint source-channel coding, and computer vision. 
He also serves as a reviewer for IEEE TVT, TAES, TCSVT, TCE, IOTJ, NeurIPS, ECCV, AAAI, and ACM MM.
\end{IEEEbiography}

\begin{IEEEbiography}[{\includegraphics[width=1in,height=1.25in,clip,keepaspectratio]{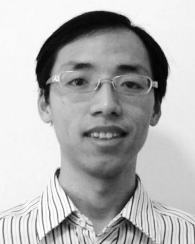}}]{Ruiqin Xiong}(Senior Member, IEEE) received the B.S. degree in computer science from the University of Science and Technology of China, Hefei, China,
in 2001, and the Ph.D. degree in computer science from the Institute of Computing Technology, Chinese Academy of Sciences, Beijing, China, in 2007.
From 2002 to 2007, he was a Research Intern with Microsoft Research Asia. From 2007 to 2009, he was a Senior Research Associate with the University of New South Wales, Sydney, NSW, Australia. In 2010, he joined the School of Electronic Engineering and Computer Science, Peking University, Beijing, where he is currently a Professor. He has authored or coauthored more than 140 technical papers in referred international journals and conferences. His research interests include image and video processing, statistical image modeling, deep learning, neuromorphic cameras, and computational imaging. He was a recipient of the Best Student Paper Award from the SPIE Conference on Visual Communications and Image Processing in 2005 and the Best Paper Award from the IEEE Visual Communications and Image Processing in 2011. He was a co-recipient  of the Best Student Paper Award from the IEEE Visual Communications and Image Processing in 2017.
\end{IEEEbiography}

\begin{IEEEbiography}[{\includegraphics[width=1in,height=1.25in,clip,keepaspectratio]{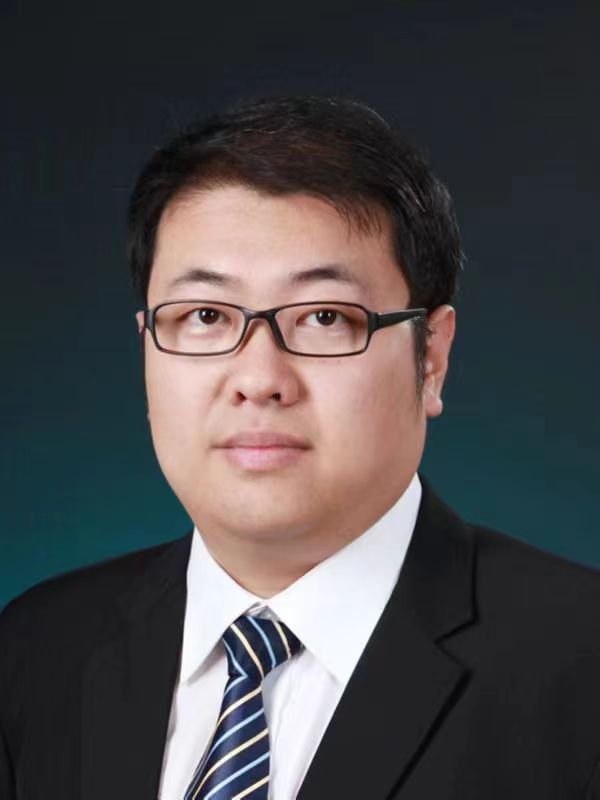}}]{Xiaopeng Fan}(Senior Member, IEEE) received the B.S. and M.S. degrees from Harbin Institute of Technology (HIT), Harbin, China, in 2001 and 2003, respectively, and the Ph.D. degree from The Hong Kong University of Science and Technology, Hong Kong, in 2009. From 2003 to 2005, he was with Intel Corporation, China, as a Software Engineer. In 2009, he joined HIT, where he is currently a Professor. From 2011 to 2012, he was with Microsoft Research Asia, as a Visiting Researcher. From 2015 to 2016, he was with The Hong Kong University of Science and Technology, as a Research Assistant Professor. He has authored one book and more than 170 papers in refereed journals and conference proceedings. His research interests include video coding and
transmission, image processing, and computer vision. He was a recipient of the Outstanding Contributions to the Development of IEEE Standard 1857 by IEEE in 2013. He was the Program Chair of PCM2017, the Chair of IEEE SGC2015, and the Co-Chair of MCSN2015. He was an Associate Editor of
IEEE 1857 Standard in 2012.
\end{IEEEbiography}

\end{document}